\documentclass[draft]{agujournal2019}
\usepackage{url} 
\usepackage{lineno}
\usepackage{amsmath}
\usepackage{amssymb}
\usepackage{natbib}
\usepackage[inline]{trackchanges} 
\usepackage{soul}
\usepackage{booktabs} 
\usepackage{array}    
\usepackage{tabularx}
\draftfalse

\journalname{Journal of Geophysical Research: Machine Learning and Computation}

\begin{document}

%
%


\title{High-Resolution Downscaling of a Lightweight Climate Emulator Using Diffusion-Based Super-Resolution}

%
%




\authors{Haiwen Guan\affil{1}, Dibyajyoti Chakraborty\affil{1}, Moein Darman\affil{2}, Troy Arcomano\affil{3}, Ashesh Chattopadhyay\affil{2}, Romit Maulik\affil{4}}

\affiliation{1}{Information Sciences and Technology, The Pennsylvania State University, University Park, Pennsylvania}
\affiliation{2}{Department of Applied Mathematics, University of California, Santa Cruz, California}
\affiliation{3}{Allen Institute for Artificial Intelligence (AI2), Seattle, Washington}
\affiliation{4}{School of Mechanical Engineering, Purdue University, West Lafayette, Indiana}





\correspondingauthor{Haiwen Guan}{hzg18@psu.edu}



\begin{keypoints}
\item We introduce a deep learning framework to downscale a coarse-grid climate emulator output from 300 km to 28 km resolution.
\item Data-driven super-resolution models can recover fine-resolution climatological statistics using inputs from a coarse-grained emulator.
\item The framework proposes an alternative to directly training climate emulators at high resolution.
\end{keypoints}

%
%

%
%
\begin{abstract}
High-resolution climate fields are essential for the assessment of their regional and local effects. Current global climate models, along with being computationally expensive, are often too coarse for applications that depend on local extremes and orographic gradients. Machine learning offers a computationally efficient route for both climate emulation and super-resolution from coarse simulations to finer spatial scales. Therefore, we develop a data-driven downscaling framework for the Lightweight Uncoupled ClImate Emulator (LUCIE), a lightweight Spherical Fourier Neural Operator climate emulator introduced by \citet{guan2025lucie}. We compare deterministic super-resolution baselines, including a U-Net and Spherical Fourier Neural Operator (SFNO), with two probabilistic diffusion formulations: a conditional Elucidated Diffusion Model and a diffusion-posterior-sampling approach. The models are trained on 6-hourly ERA5 reanalysis fields from 2000 to 2009 and evaluated on out-of-sample LUCIE predictions from 2010 to 2018, downscaling coarse emulator states to $\sim 28$~km resolution. Performance is assessed against ERA5 using latitude-weighted Root-Mean-Squared-Error (RMSE), power spectra, probability density functions, and the leading Empirical Orthogonal Function of zonal wind. These diagnostics test whether deterministic and probabilistic super-resolution models can recover fine-scale climatological statistics, high-wavenumber variability, and dominant modes with an imperfect coarse emulator. Our results characterize the strengths and limitations of the emulator-driven super-resolution, including its ability to reproduce ERA5-like fine-scale statistics while preserving explicit coarse-scale consistency with LUCIE.
\end{abstract}

\section*{Plain Language Summary}
Data-driven models are increasingly used to predict atmospheric weather and climate. While powerful, these models often face challenges such as instability over long simulations or high cost at high resolution. LUCIE is a data-driven climate emulator that produces stable, computationally efficient climate trajectories for more than 1000 years, but only at low spatial resolution---too coarse for the local climate assessments needed by communities and policymakers. In this study, we use a data-driven super-resolution framework to transform LUCIE's low-resolution outputs into high-resolution fields. We compare diffusion-based and deterministic neural-network downscalers, evaluate them rigorously against reanalysis data using climatology, spectra, and ensemble-calibration metrics, and show that this method preserves large-scale climate features while adding detailed local structure. We also document its current limitations: the downscaled ensembles are not yet well calibrated, some bias remains even with a perfect coarse input, and the underlying emulator presently produces stationary climates rather than forced future projections. The approach is therefore a promising step toward, rather than a finished tool for, high-resolution climate planning and adaptation.

\section{Introduction}

Access to high-resolution climate and weather data is critical for assessing regional risks, from hydrological modeling and energy forecasting to urban heat-stress planning. 
Further, regional-risk and climate-impact studies require locally resolved information, in addition to changes in the large-scale mean climate \citep{wan2024regional, lopez2025dynamical}.
Key impact phenomena, such as localized convective precipitation and heatwaves, are governed by fine-scale features like topography and land cover that coarse global climate products fundamentally cannot resolve \citep{SchwingshacklEtAl2024}. 
This fine-scale structure is also important for precipitation, bias correction, and kilometer-scale hazard modeling, where coarse fields smooth convective extremes, sharp gradients, and coherent mesoscale features \citep{harris2022generative, price2022increasing, aich2026conditional, hess2025fast, mardani2025residual}.

Global climate models (GCMs) remain the main tools for simulating forced climate responses under different boundary conditions and emissions scenarios. However, these exist mostly at coarse scales as
generating them at fine scale over large domains or long (decadal to centennial) periods presents a major computational bottleneck. While dynamical downscaling via regional climate models (RCMs) provides the necessary physical detail, its computational expense is prohibitive, severely limiting ensemble sizes and simulation lengths \citep{SkamarockEtAl2019, Giorgi2019}. 
Global reanalyses such as ERA5 (0.25$^\circ$) \citep{hersbach2020era5} and MERRA-2 ($0.5^\circ \times 0.625^\circ$) \citep{gelaro2017modern} can provide training and validation targets to train such weather and climate models. Models like LUCIE offer long-term, stable, and physically-consistent simulations at a fraction of the cost of traditional GCMs \citep{guan2025lucie}. 
LUCIE is part of a rapidly expanding climate emulator literature that includes ACE2, spherical probabilistic diffusion emulators, Climate in a Bottle, and coupled systems such as HiRO-ACE \citep{watt2025ace2, ruhling2024probablistic, brenowitz2025climate, perkins2025hiro}. However, their utility for impact assessment is limited by their coarse native spatial resolution, which downscaling aims to address.

Statistical and machine learning (ML) downscaling methods have emerged as a computationally efficient method for super-resolving coarse fields \citep{maraun2018statistical, rampal2024enhancing, sachindra2018statistical}.  
The field has progressed from convolutional neural networks, GANs, VAE-GANs, and neural operators \citep{vandal2017deepsd, bano2020configuration, zhu2020gan, stengel2020adversarial} to diffusion and score-based generative models that sample high-resolution fields from learned conditional distributions \citep{harris2022generative, price2022increasing, wei2023super, jiang2023efficient, TuEtAl2025_MODS, TuEtAl2025_SGD, hess2025fast, schmidt2025generative, schillinger2025enscale}. Throughout this study, we use the terms downscaling and super-resolution interchangeably. 

Recent probabilistic downscaling work has motivated several complementary modeling choices: residual-corrective diffusion recovers kilometer-scale spectra and coherent weather structures \citep{mardani2025residual}; dynamical-generative downscaling reduces the cost of climate-model ensemble downscaling \citep{lopez2025dynamical}; scale-adaptive consistency models support efficient uncertainty-aware downscaling of Earth-system fields \citep{hess2025fast}; and score-based or proper-scoring-rule approaches emphasize spatiotemporal coherence, multivariate consistency, calibration, and temporally consistent sampling \citep{schmidt2025generative, schillinger2025enscale}. Despite their success, many ML downscaling algorithms face critical limitations. Performance can degrade significantly under distributional shifts, such as when applied to geographic regions not seen during training \citep{HarderEtAl2025_RainShift}. Furthermore, a crucial and often overlooked challenge arises when models trained on ``perfect'' coarsened reanalysis data are applied to inputs from an independent climate model or emulator. 
This ``imperfect input'' setting can degrade spatial consistency and underestimate extremes when inference-time coarse forcing differs statistically from the coarsened training data \citep{ReddyEtAl2025_LimitationSR}. Moreover, some forecast-to-observation studies show that coarse-resolution source biases can materially affect generated high-resolution outputs \citep{harris2022generative, price2022increasing}. Prior work has addressed such distribution shift using explicit debiasing, domain-adaptation, and bias-correction strategies, including optimal-transport debiasing with conditional diffusion \citep{wan2023debias}, probabilistic regional-risk frameworks with model-discrepancy treatment \citep{wan2024regional}, shared embedding spaces for downscaling and bias correction \citep{aich2026conditional}, and multi-stage GCM-to-RCM generative maps with large-scale mismatch adjustment \citep{schillinger2025enscale}.

In this paper, we evaluate a framework that bridges efficient climate emulation and high-resolution impact assessment by coupling the coarse-grained LUCIE emulator with data-driven super-resolution (SR) models.
We compare deterministic baselines: a U-Net and a Spherical Fourier Neural Operator super-resolution model (SFNO-SR), along with two probabilistic diffusion formulations. The first is a conditional Elucidated Diffusion Model (EDM), following \citet{karras2022elucidating}, which directly learns to sample high-resolution states conditioned on coarse inputs. The second is a diffusion-posterior-sampling (DPS)-type approach, following \citet{chung2022diffusion} and \citet{chakraborty2026multimodal}, which combines a learned high-resolution prior with a coarse-resolution measurement operator during sampling. Conditional diffusion is specific to the purpose of super-resolving LUCIE and the DPS approach is a test-time implementation, which does not require any low-resolution data while training. In contrast to prior work that incorporates explicit debiasing, domain adaptation, or bias correction to handle the reanalysis-to-model gap \citep{wan2023debias, wan2024regional, aich2026conditional}, we deliberately retain a downscaler trained purely on reanalysis and instead evaluate how faithfully it preserves the dynamics of an independent emulator. This isolates the open question of how a generative downscaler trained only on reanalysis behaves when applied directly to an emulator, without any debiasing or domain-adaptation step. The SR task here is to resolve fine-scale details implied by the emulator's coarse state, not to fix distribution shifts. We therefore use the generative models to sample plausible high-resolution states and to diagnose how the emulator's coarse-scale biases propagate through the SR pipeline.

The SR models are trained on ERA5 reanalysis data (2000--2009) at 6-hourly cadence and subsequently evaluated on their ability to super-resolve outputs from the LUCIE emulator for a distinct period (2010--2018). The temporal resolution is chosen to match that of LUCIE to eliminate any training-validation data leakage. We evaluate the framework using a comprehensive suite of diagnostics designed to measure both point-wise accuracy and global physical consistency. In addition to standard spatial metrics, we employ zonal averaged climatology and spectral analysis to assess the models' ability to reconstruct the high-wavenumber variance and ``roughness'' of atmospheric turbulence. Furthermore, we use Empirical Orthogonal Function (EOF) analysis to verify that the SR process preserves the primary modes of climate variability, such as the annular structures of the zonal wind. We also perform robustness tests under perfect-prognosis (T30 ERA5 low resolution) and distribution shift (LUCIE) for systematic input biases.

Atmospheric SR is fundamentally an ill-posed, ``one-to-many'' problem, in which a single coarse input is physically consistent with multiple high-resolution realizations. Deterministic models such as U-Net and SFNO-SR provide a useful single-estimate accuracy and bias diagnostic, while probabilistic diffusion models generate ensembles of plausible states and enable uncertainty quantification. The remainder of this paper is organized as follows. Section~\ref{sec:methodology} describes the data sources used for training and validation, the deep-learning architectures, and the evaluation strategy. Section~\ref{sec:results} presents the experimental results, including seasonal climatology and spectral fidelity. Section~\ref{sec:discussion} discusses the broader implications of these findings and outlines future directions.

\section{Data and Methods}\label{sec:methodology}
\begin{figure}[htbp]
  \centering
  \includegraphics[width=\linewidth]{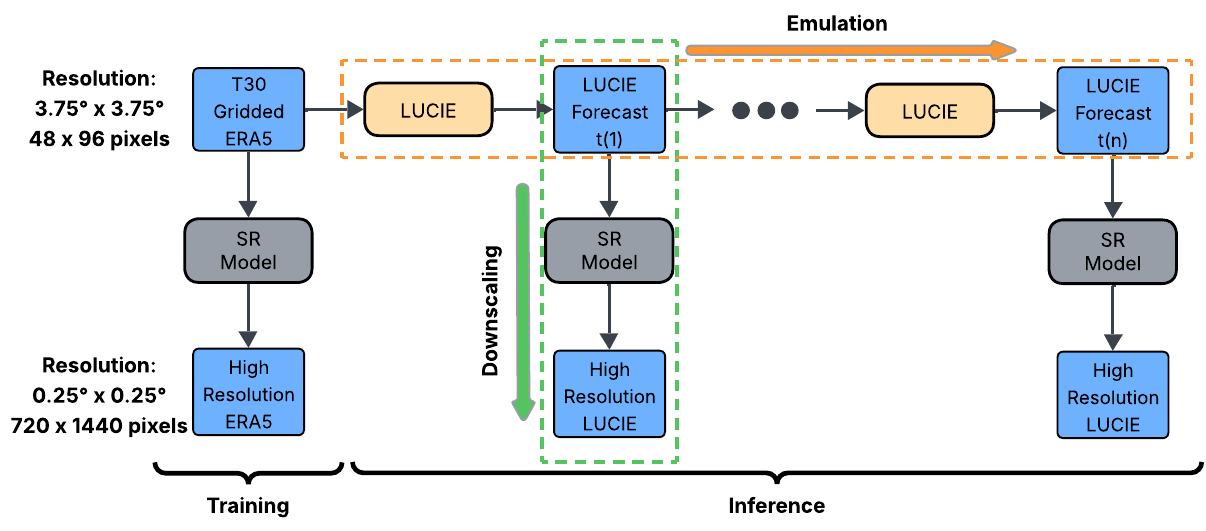}
  \caption{Schematic of the downscaling framework. \textbf{(Left) Training phase:} The downscaling model is trained on T30 gridded ERA5 data to learn the transformation from a coarse resolution to a high resolution. \textbf{(Right) Inference phase:} The trained downscaling model is applied to the coarse-resolution output from the LUCIE climate emulator to generate high-resolution climate projections.}
  \label{fig0:schematic}
\end{figure}

The primary goal of this effort is to generate high-fidelity, physically consistent climate data at a target resolution of approximately 28 km (720×1440 gridpoints/pixels), given predictions from a coarse-grained emulator at a resolution of approximately 300 km ($48 \times 96$). To accomplish this, we construct a coupling scenario as shown in Figure~\ref{fig0:schematic}. A coarse-grained atmospheric forecast is initialized using LUCIE on a T30 Gridded ERA5 state at a timestep of 6 hours. Subsequently, the downscaling model is applied as a postprocessing step at each timestep of the prediction. The fine scale state is \emph{not} reused to initialize the emulator for the next step prediction implying that this downscaling operation can be considered to be a postprocessing approach that can be performed offline. We detail the various components of this approach in the following subsections. 


\subsection{LUCIE}

The coarse-grid dynamics that are used as inputs to the downscaling framework are provided by the LUCIE emulator \citep{guan2025lucie}. LUCIE (Lightweight Uncoupled ClImate Emulator) is a data-driven atmospheric emulator designed to overcome the prohibitive computational costs of traditional numerical weather prediction and high-resolution ML models, which often require thousands of GPU hours. By leveraging an SFNO \citep{bonev2023spherical} backbone and a specialized training strategy—including spectral regularization and a hard-constrained first-order integrator—LUCIE maintains long-term stability and physical consistency without the common unphysical drift in autoregressive models. It is trained on six-hourly ERA5 at T30 gaussian grid with six prognostic and diagnostic variables, and can be run for hundreds of years with large ensembles.

The primary advantage of LUCIE lies in its computational efficiency; it can be trained in just 2.4 hours on a single A100 GPU and is capable of simulating 6,000 years of climate per day. However, the relatively coarse T30 Gaussian grid ($\sim 3.75^{\circ}$ resolution) is its main limitation. While this grid is sufficient for capturing large-scale dynamics and global climatology, it cannot resolve the fine-scale regional details necessary for localized studies.

As our framework downscales LUCIE rather than correcting it, LUCIE's own biases are inherited by the high resolution output, and we therefore summarize the most relevant ones reported by \citet{guan2025lucie}. LUCIE reproduces the long-term climatology and variability well, but exhibits a near-surface warm bias over land (global RMSE $\approx 0.727$~K, mostly bounded within $\pm 5$~K), places the Northern Hemisphere jet slightly too far poleward with an overactive subtropical jet, and underpredicts annual precipitation, most notably over the Maritime Continent. It captures the Southern Annular Mode accurately but represents the Northern Annular Mode less faithfully, and it tends to overestimate the frequency of extremes for most variables. These biases are relevant context for the downscaling results that follow, since they set a floor on the fidelity achievable when LUCIE is used as input rather than reanalysis.

This study proposes a downscaling framework as a critical bridge for this gap. By coupling LUCIE’s stable, low-cost dynamical core with machine learning downscaling, we overcome the T30 resolution barrier and extend LUCIE’s utility to regional-scale applications. This two-stage approach leverages the complementary strengths of both models: the computational efficiency of the lightweight emulator and the high-fidelity spatial reconstruction of the diffusion framework, enabling detailed analysis of local phenomena without the computational cost of end-to-end high-resolution training.

\subsection{Data}

For training our downscaling models, we use ECMWF Reanalysis v5 (ERA5) \citep{hersbach2020era5}, a physically consistent, global atmospheric reanalysis available hourly on a $0.25^{\circ}$ grid. ERA5 combines ECMWF's Integrated Forecasting System (IFS) with data assimilation of both conventional measurements and a wide range of satellite observations, providing a long, temporally consistent record from the mid-20th century to the present. For our SR experiments, the low-resolution inputs are ERA5 fields regridded to a T30 Gaussian grid ($\approx3.75^{\circ}$) \citep{arcomano2022hybrid}, while the high-resolution targets are the native $0.25^{\circ}$ ERA5 fields. This pairing is chosen to enable zero-shot inference of the trained SR model with the LUCIE climate emulator, which itself operates on T30 Gaussian grid ERA5. To match the emulator’s channel set, we include near-surface temperature (model level 133, $\sigma\approx0.95$, $\approx 963$ hPa over ocean), zonal and meridional winds at model level 83 ($\sigma\approx0.34$, $\approx 345$ hPa over ocean), and surface precipitation. The training dataset is set to cover the year 2000 to year 2009 with samples every 6 hours. The data is normalized by mean and standard deviation over the training period. For the deterministic models SFNO-SR and UNet-SR in this study, the input data is bicubic interpolated from $3.75^{\circ}$ to $0.25^{\circ}$. All the models have 0.25$^\circ$ orography as the forcing variable (input only). The variables are listed in Table~\ref{tab:data_variables}.

\begin{table}[ht]
\centering
\caption{The list of low-resolution input variables, high-resolution target variables, the forcing variable, and their grid resolution. Model level 133 corresponds to $\sigma\approx0.95$ ($\approx 963$~hPa) and model level 83 corresponds to $\sigma\approx0.34$ ($\approx 345$~hPa).}
\label{tab:data_variables}
\begin{tabular}{@{}lll@{}}
\toprule
\textbf{Category} & \textbf{Variables} & \textbf{Resolution} \\ \midrule
\textbf{Input (Low-Res)} & \begin{tabular}[c]{@{}l@{}}Temperature (model level 133, $\sigma\approx0.95$)\\ Zonal Wind ($u$, model level 83, $\sigma\approx0.34$)\\ Meridional Wind ($v$, model level 83, $\sigma\approx0.34$)\\ Surface Precipitation\end{tabular} & \begin{tabular}[c]{@{}l@{}}T30 Gaussian\\ ($\approx 3.75^{\circ}$)\end{tabular} \\ \midrule
\textbf{Target (High-Res)} & \begin{tabular}[c]{@{}l@{}}Temperature (model level 133, $\sigma\approx0.95$)\\ Zonal Wind ($u$, model level 83, $\sigma\approx0.34$)\\ Meridional Wind ($v$, model level 83, $\sigma\approx0.34$)\\ Surface Precipitation\end{tabular} & \begin{tabular}[c]{@{}l@{}}ERA5 Native\\ ($0.25^{\circ}$)\end{tabular} \\ \midrule
\textbf{Forcing (Static)} & Orography & $0.25^{\circ}$ \\ \midrule
\end{tabular}
\end{table}

\subsection{SFNO-SR}
Neural operators represent a machine learning framework that may be used to learn transformations between continuous function spaces. The Fourier Neural Operator (FNO) \citep{li2020fourier} implements this by performing convolutions in the spectral domain, which allows for global spatial reasoning and inherent resolution-invariance. To account for the Earth's geometry, the SFNO replaces standard Fourier Transforms with Spherical Harmonic Transforms (SHT) \citep{bonev2023spherical}. This ensures that the model respects the spherical coordinate system, avoiding the artifacts around boundaries that arise when processing atmospheric data on a flattened projection. This architecture is better suited for the data-driven modeling task as compared to the other UNet-type models since it respects the spherical geometry of the data.

The SFNO-SR architecture leverages this framework to map four variables—temperature, zonal wind ($u$), meridional wind ($v$), and precipitation—from a $48 \times 96$ grid to a $720 \times 1440$ resolution ($15\times$ finer along each axis in grid-count terms, corresponding to the $\sim 300$~km $\rightarrow \sim 28$~km change in nominal resolution). Additionally, it uses orography as an input only condition. The process projects inputs into a $32$-dimensional latent space, followed by four SFNO layers: an encoder, a spectral upscaling layer, and two high-resolution blocks \citep{bonev2023spherical}. These layers utilize SHTs and complex-valued kernels for spatial operations, interspersed with four 8-head self-attention blocks to capture long-range dependencies. A long skip connection connects the initial encoder directly to the final processing blocks. The model uses Gaussian Error Linear Unit (GELU) activations and instance normalization throughout. Training was conducted over $100$ epochs using a Mean Absolute Error ($L_1$) loss function and the Adam optimizer with a learning rate of $0.001$, following a cosine annealing schedule. We chose $L_1$ rather than mean squared error because squared-error training is dominated by rare large-magnitude precipitation events and tends to produce excessively smoothed precipitation fields, an effect documented in earlier ML downscaling work \citep{harris2022generative, price2022increasing}; the resulting models therefore target the conditional median, while RMSE in Table~\ref{tab:rmse} is reported for direct comparability with prior downscaling studies that use it as the standard metric. This architecture is chosen so that SFNO-SR shares the operator backbone with the LUCIE climate emulator. As shown in the following sections, SFNO-SR is not the best-performing model in our experiments; its purpose here is to establish a deterministic baseline whose backbone matches LUCIE, so that the downscaler can be embedded directly inside the emulator in a future end-to-end SFNO emulation-and-downscaling pipeline.

\subsection{UNet-SR}
The UNet \citep{ronneberger2015u} is a specialized convolutional neural network architecture characterized by a symmetrical, U-shaped design that utilizes skip connections to bridge high-resolution features from an encoder path to a decoder path for spatial reconstruction. Building on its success in image processing, the UNet architecture has been widely adapted for super-resolution within weather forecasting and atmospheric science \citep{sharma2022resdeepd, zhang2024super}. We utilize the standard encoder-decoder UNet framework \citep{ronneberger2015u} as our baseline UNet-SR model to perform spatial super-resolution on global climate data. The model takes five input channels: the four primary weather variables (temperature, $u$-wind, $v$-wind, and precipitation) and the static orography (surface geopotential) field as an input-only forcing variable, the same forcing used by the other models (Table~\ref{tab:data_variables}). Prior to entering the network, the low-resolution T30 data is upsampled to the target $720 \times 1440$ resolution using bicubic interpolation, and precipitation values are normalized via a log-scaling transformation. The encoder path consists of four stages of downsampling, using $2 \times 2$ max-pooling and double $3 \times 3$ convolutions to increase the feature depth from 16 to a 256-channel bottleneck. The decoder path mirrors this structure, employing bilinear interpolation for upsampling and direct concatenation skip connections at each level to recover spatial details from the encoder. Each convolution is followed by Batch Normalization and ReLU activation. The model was trained for 100 epochs using Mean Absolute Error ($L_1$) loss and the Adam optimizer with a cosine annealing learning rate schedule.

\subsection{Elucidating Diffusion Models (EDM)}
\label{subsec:edm_basics}
We approach the generative modeling task using the diffusion framework, which models data generation as a reversal of a progressive noise corruption process \citep{sohl2015deep, ho2020denoising, song2020score}. Specifically, we adopt the Elucidating Diffusion Model (EDM) formulation proposed by \citet{karras2022elucidating}, which provides a robust design space for score-based generative models. We perform conditional SR via input concatenation, and finally utilize an unconditional model with a posterior sampling strategy.

The forward process in the EDM framework describes the gradual corruption of a clean data sample $\mathbf{x}_0 \sim p_{\text{data}}(\mathbf{x})$ by additive Gaussian noise.
\begin{equation}
    \mathbf{x}(\sigma) = \mathbf{x}_0 + \sigma \mathbf{n}, \quad \text{where} \quad \mathbf{n} \sim \mathcal{N}(\mathbf{0}, \mathbf{I}).
\end{equation}
The generative process involves solving a probability flow Ordinary Differential Equation (ODE) that traverses from a high-noise distribution ($\sigma = \sigma_{\max}$) back to the data distribution ($\sigma = \sigma_{\min}$) (see Appendix~\ref{hyperparams} for details on these). The ODE is defined as:
\begin{equation}\label{PFODE}
    d\mathbf{x} = \left[ \frac{\mathbf{x}(\sigma) - D_\theta(\mathbf{x}(\sigma);\sigma)}{\sigma} \right] d\sigma,
\end{equation}
where $D_\theta(\mathbf{x}(\sigma);\sigma)$ is the learned denoiser's estimate of the clean data. The bracketed term equals $-\sigma\,\nabla_{\mathbf{x}}\log p(\mathbf{x};\sigma)$, i.e., the (negative) score of the noised distribution scaled by $\sigma$, since for the Gaussian perturbation kernel the score satisfies $\nabla_{\mathbf{x}}\log p(\mathbf{x};\sigma) = (D_\theta(\mathbf{x};\sigma)-\mathbf{x})/\sigma^2$. At sampling time the true clean state $\mathbf{x}_0$ is unavailable, so it is replaced by the denoiser estimate $D_\theta$. To this end, a neural network $F_\theta$ is trained to estimate the clean data $\mathbf{x}_0$ from a noisy input. EDM introduces a preconditioning scheme to ensure the network inputs and outputs maintain unit variance during training. The effective denoiser $D_\theta$ is trained by minimizing the weighted $L_2$ error between the predicted and ground-truth data:
\begin{equation}
    \mathcal{L}(\theta) = \mathbb{E}_{\mathbf{x}_0, \mathbf{n}, \sigma} \left[ \lambda(\sigma) \| D_{\theta}(\mathbf{x}_0 + \sigma \mathbf{n}; \sigma) - \mathbf{x}_0 \|_2^2 \right],
    \label{eq:edm_loss}
\end{equation}
where the weighting function is typically set to weight inputs at different noise levels differently. See \citet{karras2022elucidating} for more details on these hyperparameters. Once the model is trained to predict the expected $\mathbf{x}_0$ at different noise levels, it is substituted in Equation~\ref{PFODE} and integrated to obtain samples from the data distribution. We use the stochastic sampler mentioned in \citet{karras2022elucidating} and \citet{chakraborty2026multimodal}.

\subsubsection{Conditional Downscaling}
\label{subsec:conditional_sr}

To adapt the EDM framework for SR, we modify the network to estimate the high-resolution field $\mathbf{x}_0$ conditioned on a low-resolution field $\mathbf{y}$. A straightforward and effective method for incorporating this condition is channel-wise concatenation of a bicubically upsampled version of $\mathbf{y}$ with the noisy high-resolution state \citep{mardani2024residual}.

Let $\mathcal{U}(\cdot)$ denote an upsampling operator (e.g., bicubic interpolation) that projects $\mathbf{y}$ to the spatial resolution of $\mathbf{x}$. The neural network input is augmented to include this guidance information:
\begin{equation}
    \mathbf{x}_{\text{in}} = \text{Concat}(\mathbf{x}(\sigma), \mathcal{U}(\mathbf{y})).
\end{equation}
The denoiser becomes a conditional function $D_{\theta}(\mathbf{x}(\sigma), \mathbf{y}; \sigma)$. By providing the low-resolution context directly to the early layers of the model, the network learns to synthesize high-frequency details that are statistically consistent with the coarse-grained structures present in $\mathbf{y}$. In effect, the conditional denoiser learns the conditional distribution $p(\mathbf{x}_0\mid\mathbf{y})$ directly: its training objective is the conditional form of Equation~\ref{eq:edm_loss}, in which the loss is evaluated on samples drawn from the joint distribution of high- and low-resolution pairs. The low-resolution constraint is thus absorbed into the trained network rather than imposed at sampling time, which contrasts with the posterior-sampling approach of Section~\ref{subsec:posterior_sampling}.

The conditional super-resolution framework utilizes a U-Net backbone optimized with EDM (Elucidating Diffusion Models) preconditioning. The U-Net is modified to take $\sigma$ as an extra conditioning variable \citep{song2020score}. The architecture is designed for multi-variable atmospheric states, featuring 4 input/output channels (temperature, $u$-wind, $v$-wind, and precipitation) conditioned on the 4 coarse-resolution T30 channels and orography from high-resolution ERA5. To capture multi-scale atmospheric features without the computational overhead of self-attention, we employ a deep residual structure with 3 residual blocks per level and a progressive channel multiplier sequence of [1, 2, 4, 8, 16], starting from a base width of 32 channels. Group Normalization (32 groups) is applied throughout to maintain training stability across varying noise scales. Further details of common hyperparameters are provided in Appendix~\ref{hyperparams}.

\subsubsection{Posterior Sampling Downscaling}
\label{subsec:posterior_sampling}

The limitation of our previous approach is that a new model needs to be trained every time the low-resolution data source is changed. Whereas the conditional model of Section~\ref{subsec:conditional_sr} learns $p(\mathbf{x}\mid\mathbf{y})$ directly, posterior sampling instead uses only the learned unconditional prior $p(\mathbf{x})$, represented by the denoiser $D_\theta$, and incorporates the coarse observation through a likelihood term applied at sampling time. Because the prior is independent of the observation operator, the same trained model can be reused with a different coarse data source or measurement operator without retraining. To generate samples consistent with different partial or low-fidelity observations $\mathbf{y}$, we employ a posterior sampling strategy \citep{chakraborty2026multimodal,schmidt2025generative}. By applying Bayes' rule, the score function of the posterior distribution $p(\mathbf{x}|\mathbf{y})$ decomposes into the learned prior score and a likelihood constraint:
\begin{equation}
    \nabla_{\mathbf{x}} \log p(\mathbf{x}|\mathbf{y}) = \nabla_{\mathbf{x}} \log p(\mathbf{x}) + \nabla_{\mathbf{x}} \log p(\mathbf{y}|\mathbf{x})
    \label{eq:posterior_split}
\end{equation}
This decomposition is the basis of the DPS approach, and it is distinct from the conditional model of Section~\ref{subsec:conditional_sr}: the conditional model learns the posterior score $\nabla_{\mathbf{x}}\log p(\mathbf{x}|\mathbf{y})$ directly through training on paired data, whereas DPS never trains a conditional model and instead assembles the posterior score at sampling time from a separately learned unconditional prior $\nabla_{\mathbf{x}}\log p(\mathbf{x})$ and an explicit likelihood term $\nabla_{\mathbf{x}}\log p(\mathbf{y}|\mathbf{x})$. Evaluating the likelihood $p(\mathbf{y}|\mathbf{x}(\sigma))$ on noisy states is analytically intractable. Following methods like Diffusion Posterior Sampling (DPS) \citep{chung2022diffusion} and Score-based Data Assimilation (SDA) \citep{rozet2023score}, we approximate this term using the denoised estimate of the data. Since the EDM denoiser directly predicts the clean data, i.e., $\hat{\mathbf{x}}_0 = D_\theta(\mathbf{x}(\sigma); \sigma)$, we can compute a likelihood score $\mathbf{s}_l$ by backpropagating through the differentiable measurement operator $\mathcal{M}$:
\begin{equation}
    \mathbf{s}_l(\mathbf{x}, \sigma; \mathbf{y}) \approx - \nabla_{\mathbf{x}} \frac{||\mathbf{y} - \mathcal{M}(D_\theta(\mathbf{x}; \sigma))||^2}{\Sigma_y + \sigma^2 \hat{\Gamma}}
    \label{eq:likelihood_score_edm}
\end{equation}
The denominator represents the effective variance, combining the measurement noise covariance $\Sigma_y$ with a time-dependent stability term $\sigma^2 \hat{\Gamma}$ derived from the SDA formulation \citep{rozet2023score}, where $\hat{\Gamma}$ is a tunable constant approximating the Jacobian-prior interaction. The measurement operator $\mathcal{M}$ maps a high-resolution field to a lower resolution by simple bicubic interpolation, so that the residual $\mathbf{y}-\mathcal{M}(D_\theta(\mathbf{x};\sigma))$ in Equation~\ref{eq:likelihood_score_edm} is evaluated at the coarse resolution of $\mathbf{y}$. We deliberately use a generic differentiable bicubic downsampling rather than the exact regridding that produced the T30 Gaussian-grid training data. Tying the likelihood to a single, dataset-specific coarsening operator would make the method less general, whereas a plain bicubic interpolation lets the same prior be reused with different low-resolution sources without redefining $\mathcal{M}$. This makes $\mathcal{M}$ only an approximation of the exact coarsening that brings high-resolution ERA5 to the low-resolution grid of the LUCIE training data, and we empirically adjust the measurement noise to absorb the resulting mismatch.

The constant $\hat{\Gamma}$ and the assumed measurement-noise level together set the relative weighting of the prior and the likelihood in Equation~\ref{eq:likelihood_score_edm}, and therefore control both how closely DPS follows its coarse conditioning and how much its ensemble members spread. A larger effective variance down-weights the likelihood, so samples are drawn more freely from the prior; a smaller one pulls every member toward the same coarse measurement and narrows the ensemble. The configuration used here yields an under-dispersed ensemble whose members are mutually similar, consistent with the spread-skill ratios reported in Section~\ref{sec:calibration} (Table~\ref{tab:ssr}). We did not perform a systematic sweep over $\hat{\Gamma}$ and the measurement-noise level; a study of how this weighting trades off conditioning fidelity against ensemble spread is a natural direction for improving DPS calibration, albeit at significantly increased computational cost, which we leave for future work.

The DPS model employs the same UNet with EDM preconditioning as the conditional SR model, maintaining 3 residual blocks per level and a channel multiplier sequence of [1, 2, 4, 8, 16]. A critical distinction here is that the diffusion model is not conditioned on any low-resolution inputs. The likelihood score term is used only during sampling. This configuration enables the model to serve as a learned prior for atmospheric states, where the reverse diffusion process is guided by the coarse input to satisfy the posterior distribution $p(x_0|y)$. The likelihood score term guides the generation process by modifying the probability flow ODE to drift towards regions of high likelihood. At each discretization step, the standard EDM drift is adjusted to project the trajectory toward the measurement manifold. The full procedure is detailed in Algorithm 1 in \citet{chakraborty2026multimodal}. Details of common hyperparameters are provided in Appendix~\ref{hyperparams}.

\subsection{Computational Cost}
\label{sec:cost}

All training was performed on NVIDIA A100 (80GB) GPUs using distributed data parallel training. Both the Conditional EDM and the DPS framework required approximately 23 hours of training. For inference, the Conditional EDM generates a 10-year high-resolution dataset (used here as a throughput benchmark; the climatological evaluation uses the 2010--2018 subset) in 3.5 hours on a single GPU ($\approx$ 68.5 years/GPU-day). In contrast, DPS—which requires iterative gradient-based guidance—demands significantly higher resources, requiring 16 GPUs to generate a 10-year dataset in 4 hours ($\approx$ 3.75 years/GPU-day). While the SFNO-SR baseline provides high spectral fidelity, the deterministic UNet is the most computationally efficient model in our study. The UNet requires only 0.1 seconds for a single 6-hourly inference step, achieving a throughput of 591.7 years/GPU-day. In comparison, the SFNO-SR is more complex, requiring 12.5 GPU-hours for training and 0.68 seconds per sample (approximately 86.8 years/GPU-day) for inference on a single A100 GPU.

\section{Results}\label{sec:results}

This section evaluates the performance of the proposed super-resolution (SR) frameworks in downscaling global climate data, using the LUCIE model’s low-resolution outputs as a baseline. We assess the models across three critical aspects of atmospheric fidelity: climatological accuracy, where we examine long-term means and regional spatial fidelity; statistical consistency, using power spectra and probability density functions (PDFs) to verify the preservation of multiscale variability; and climate variability, through an EOF analysis of primary modes of atmospheric circulation. We compare two families of methods throughout: \emph{deterministic} baselines (Bicubic, SFNO-SR, UNet-SR), denoted with a dagger ($\dagger$), and \emph{generative} diffusion downscalers (Conditional EDM and Diffusion Posterior Sampling, hereafter DPS), denoted with an asterisk ($\ast$). The two families are kept visually separable: in Table~\ref{tab:rmse} the deterministic and generative methods are grouped under explicit column headers and tagged inline by $\dagger$ and $\ast$; the calibration tables (Tables~\ref{tab:crps}, \ref{tab:ssr}) report only the generative methods, which are designed to produce a calibrated ensemble, whereas the deterministic baselines are single-point estimates assessed by RMSE; and figures consistently draw the generative methods as ensemble-mean curves with shaded ensemble-spread bands while the deterministic baselines are drawn as solid lines without shading. Some figure panels retain the original method labels ``conditional'' and ``Posterior Sampling''; these denote the Conditional EDM and DPS, respectively. We evaluate both families on point-wise accuracy, spectral and distributional fidelity, and ensemble calibration.

\subsection{Climatology}
\label{sec:climatology}

\begin{figure}[h!]
  \centering
  \includegraphics[width=\linewidth]{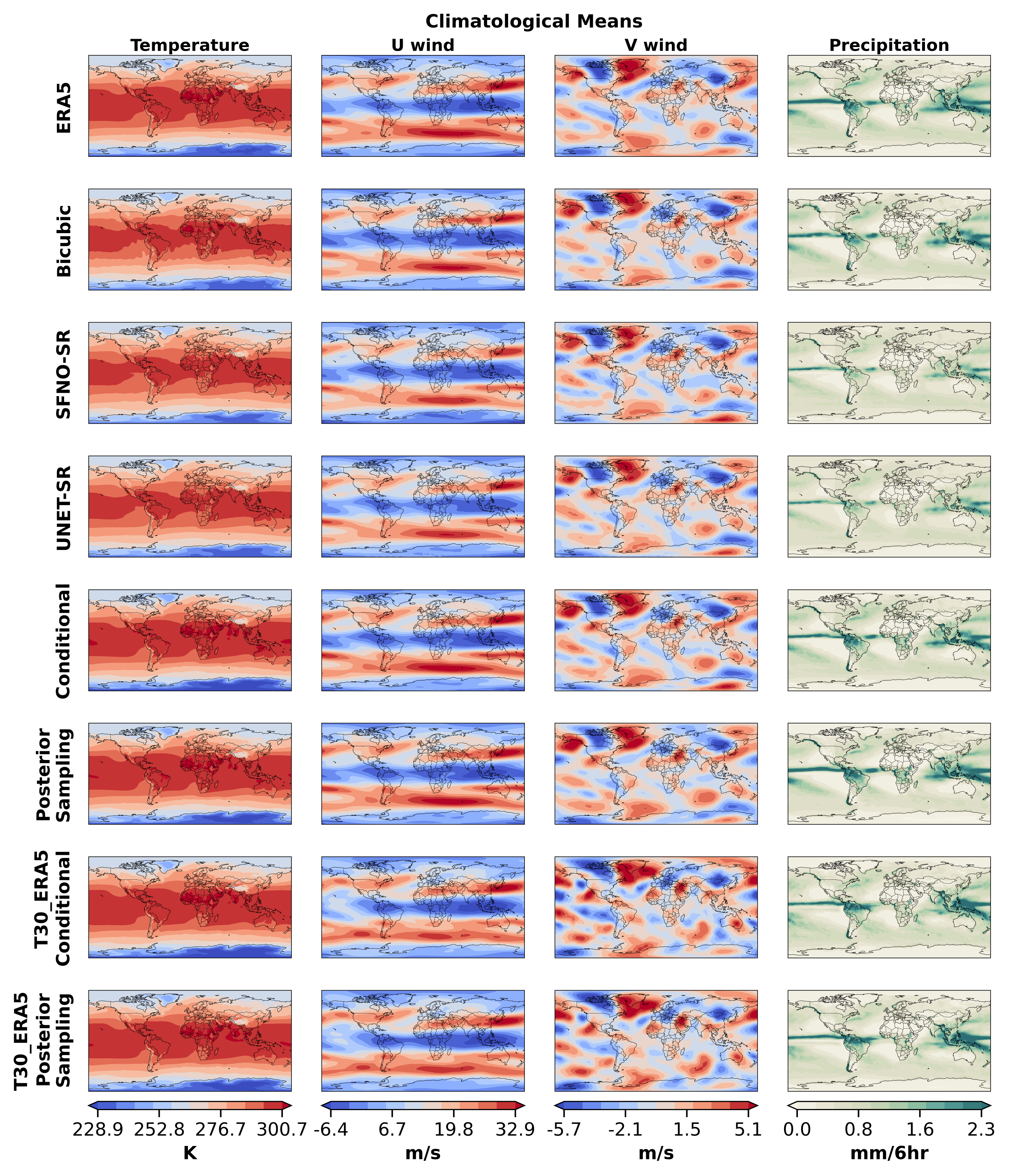}
  \caption{Super-resolved climatological snapshots averaged over the 2010--2018 period for near-surface temperature, zonal wind, meridional wind, and precipitation. Coarse-grid dynamics for this period were generated by the LUCIE emulator and then downscaled with different super-resolution algorithms corresponding to the different rows. ERA5 reanalysis (our assumed ground truth) is provided in the first row from the top. The last two rows (labelled T30\_ERA5) show the perfect-prognosis sampling experiments, in which the diffusion models downscale ERA5 coarsened to T30 instead of LUCIE output.}
  \label{fig1:climatology}
\end{figure}

As a standard and informative indicator of climatological performance, we begin by evaluating the long-term climatological means of super-resolved near-surface temperature, zonal wind at model level 83 ($\sigma\approx0.34$, $\approx 345$~hPa), meridional wind at model level 83 ($\sigma\approx0.34$, $\approx 345$~hPa), and surface precipitation over the nine-year LUCIE inference period 2010--2018. Figure~\ref{fig1:climatology} presents these climatological fields for ERA5, bicubic interpolation, SFNO-SR, UNet-SR, the Conditional EDM, and DPS. As expected for purely statistical downscaling, bicubic interpolation produces the major pattern in the climatological means but is missing fine details in the regional areas. The diffusion-based SR models generate high-resolution fields with accurate long-term average compared to UNet-SR and SFNO-SR, as reflected in the RMSE values summarized in Table~\ref{tab:rmse}: the Conditional EDM reduces the errors substantially for temperature, precipitation, and zonal wind (for example, temperature from 2.427~K for UNet-SR to 1.357~K), while for the meridional wind the improvement is marginal---the Conditional EDM (0.770~m\,s$^{-1}$) only slightly beats the best deterministic baseline, SFNO-SR (0.780~m\,s$^{-1}$). DPS achieves the lowest RMSE in temperature and zonal wind while maintaining competitive performance for meridional wind and precipitation. While deterministic baselines like UNet-SR achieve competitive RMSE, they produce visibly oversmoothed fields. The STD reported in Table~\ref{tab:rmse} is the temporal standard deviation of each field about its own climatological mean---a measure of how much climatological variability the model carries in time---and is a distinct quantity from the ensemble spread of the generative models, which we assess separately through the spread-skill ratio in Section~\ref{sec:calibration}. The two should not be conflated: the temporal STD here characterizes a single field's variability over the record, whereas the spread-skill ratio characterizes the dispersion of an ensemble at a fixed time. The deterministic UNet-SR has a lower temporal STD (4.91~K for temperature) than the diffusion-based models, consistent with the oversmoothing visible in the spectra (Figure~\ref{fig3:metrics}, where the deterministic baselines lose the high-wavenumber tail) and in the regional maps (Figures~\ref{fig:DJF}--\ref{fig:MAM}). We do not interpret the raw STD as ``larger is better''---without an ERA5 reference STD on the same grid we cannot say which model is closest to the truth, and a higher STD could in principle indicate over-dispersion---but the spectra and maps independently confirm that the deterministic baselines under-represent fine-scale variability. However, despite the visual performance, generative downscaling introduces ensemble variability and resolves mesoscale structure that is absent from the coarse emulator, which inflates the point-wise RMSE relative to the coarse emulator's own evaluation. For example, the DPS temperature field has an RMSE of 1.237~K against ERA5 at $0.25^\circ$, compared with 0.727~K for native LUCIE temperature evaluated against ERA5 on the T30 grid \citep{guan2025lucie}. These two numbers are not a like-for-like comparison---they are evaluated at very different resolutions and the high-resolution evaluation is penalized by fine-scale variability that the T30 evaluation averages out---but the contrast illustrates the variance penalty that any generative downscaler pays for restoring fine-scale structure.

\begin{figure}[h!]
  \centering
  \includegraphics[width=\linewidth]{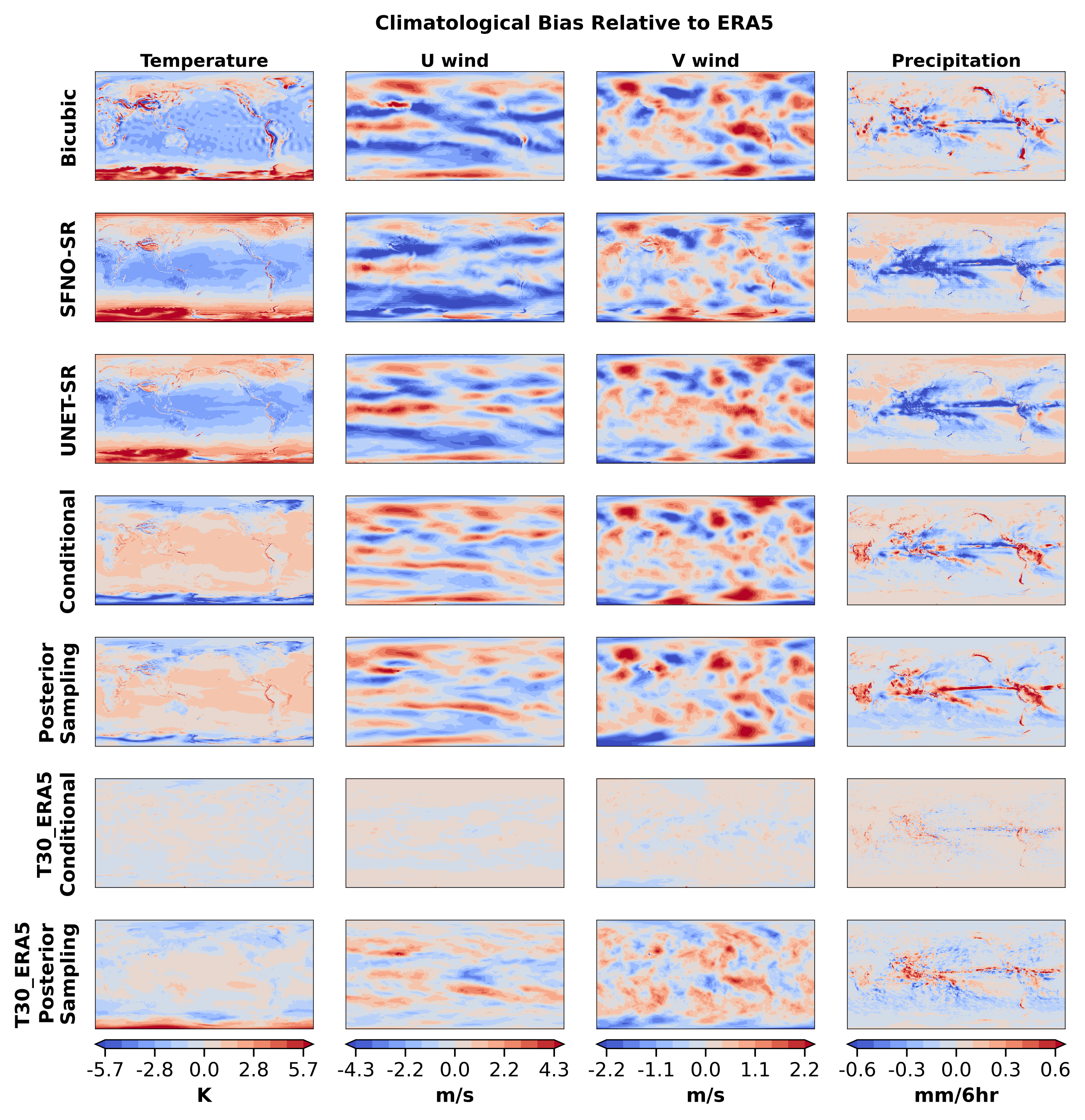}
  \caption{Climatological bias maps for each downscaling model result compared to ERA5. Each panel shows the model climatological mean minus the ERA5 climatological mean for the corresponding variable. Rows show deterministic baselines, generative diffusion downscalers applied to LUCIE outputs, and T30\_ERA5-conditioned diffusion downscalers. The T30\_ERA5-conditioned rows are evaluated against the matching ERA5 subset used for the perfect-prognosis experiments.}
  \label{fig1b:climatology_bias}
\end{figure}

\begin{figure}[h]
  \centering
  \includegraphics[width=\linewidth]{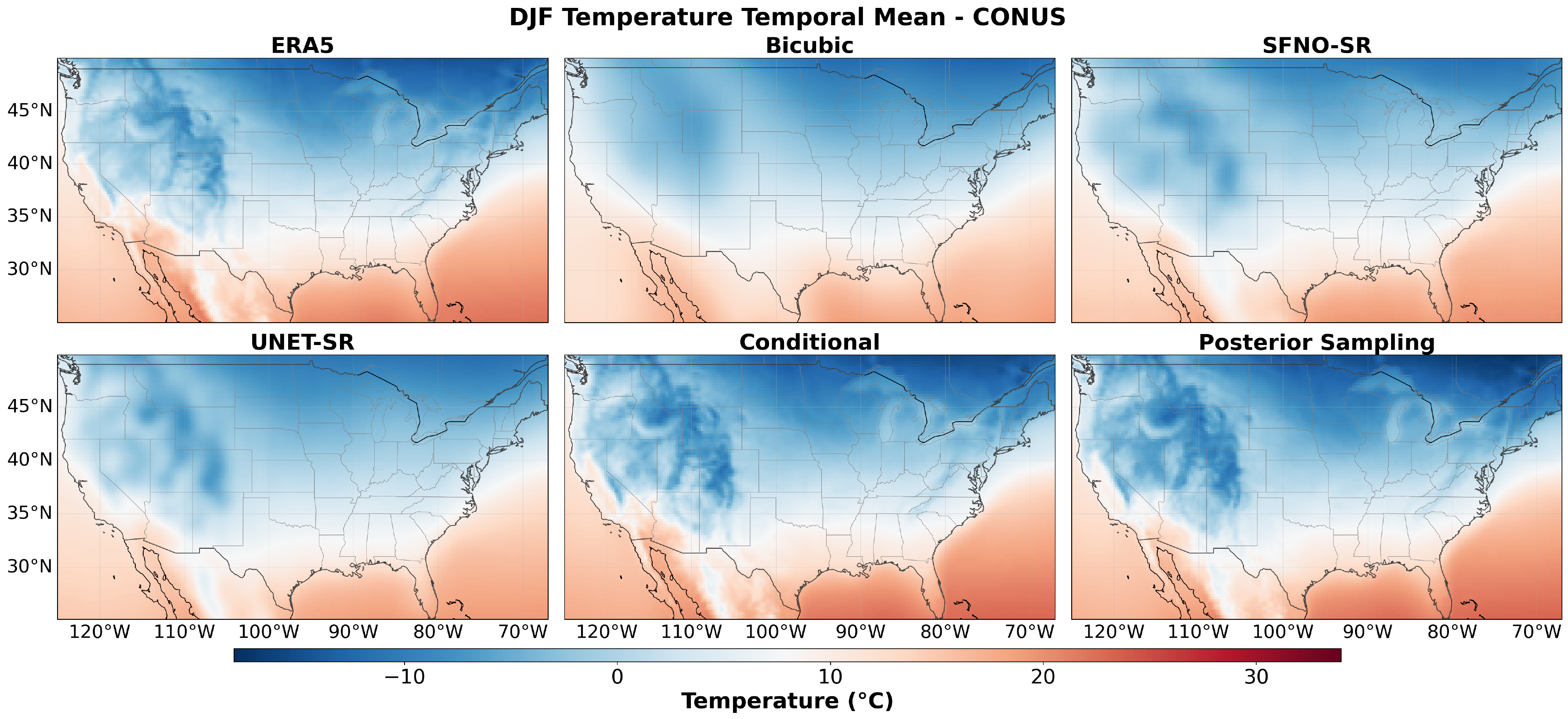}
  \caption{Climatological-mean near-surface temperature ($^\circ$C) over the Contiguous United States (CONUS) during boreal winter (December--January--February, DJF), averaged over 2010--2018. Panels show ERA5 (top left, taken as ground truth), bicubic interpolation, SFNO-SR, UNet-SR, the Conditional EDM, and DPS, each applied to LUCIE output. The deterministic baselines (bicubic, SFNO-SR, UNet-SR) produce a smoothed temperature field, whereas the diffusion-based models (Conditional EDM and DPS) recover the sharp thermal gradients and cold-air structures over the Intermountain West and Northern Plains seen in ERA5. The color scale is shared across all panels.}
  \label{fig:DJF}
\end{figure}

\begin{figure}[h]
  \centering
  \includegraphics[width=\linewidth]{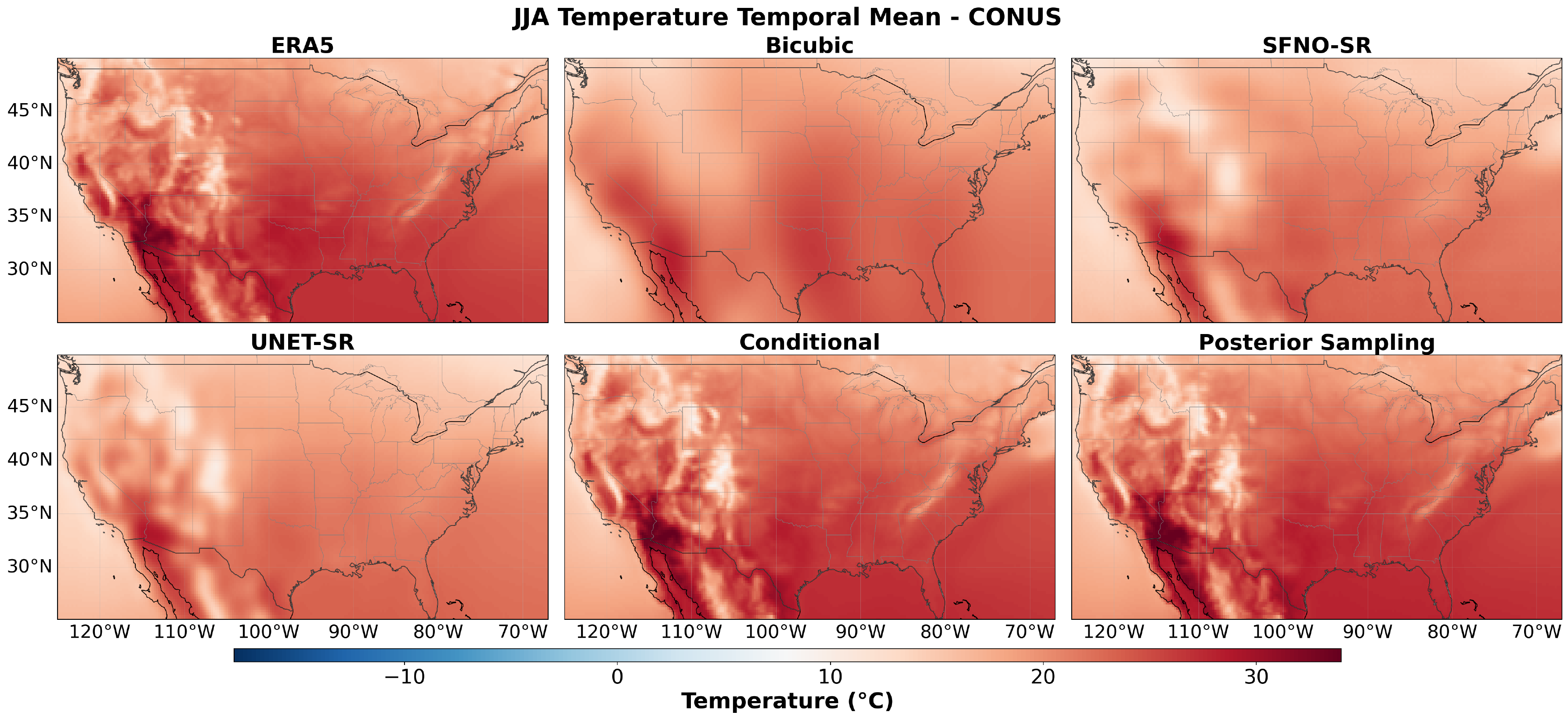}
  \caption{Climatological-mean near-surface temperature ($^\circ$C) over the CONUS during boreal summer (June--July--August, JJA), averaged over 2010--2018. Panels show ERA5 (top left, taken as ground truth), bicubic interpolation, SFNO-SR, UNet-SR, the Conditional EDM, and DPS, each applied to LUCIE output. The deterministic baselines provide only a generalized heat distribution and miss the cool high altitude ``islands'' associated with the Rocky Mountains and Sierra Nevada, whereas the diffusion-based models resolve these fine-scale orographic structures and the thermal contrast of the Central Valley of California, consistent with ERA5. Color scale is shared across all panels.}
  \label{fig:JJA}
\end{figure}

\begin{figure}[h]
  \centering
  \includegraphics[width=\linewidth]{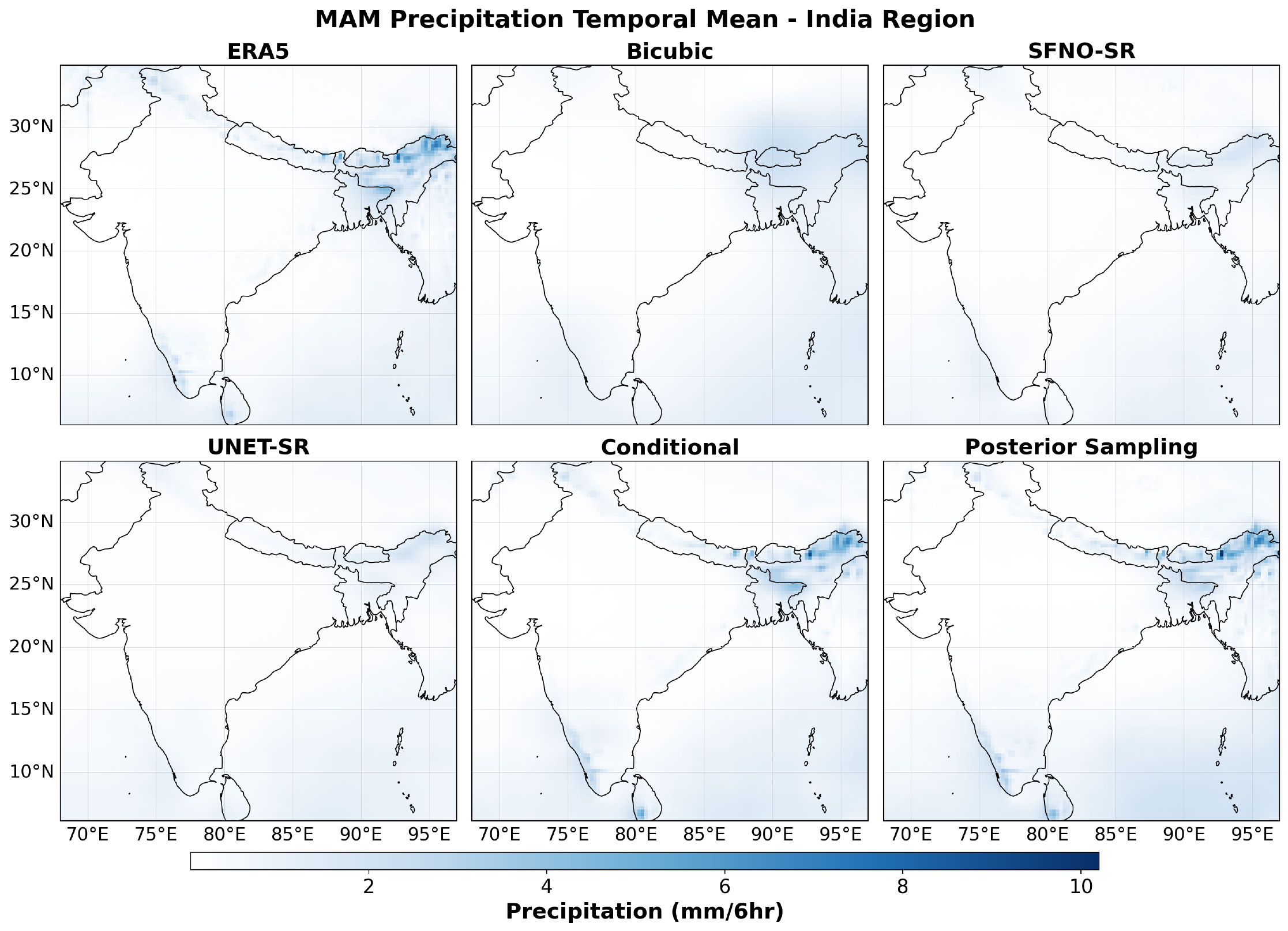}
  \caption{Temporal-mean precipitation (mm/6hr) over the India region during the pre-monsoon season (March--April--May, MAM), averaged over 2010--2018. Panels compare ERA5 (top left, taken as ground truth) against bicubic interpolation, SFNO-SR, UNet-SR, the Conditional EDM, and DPS, each applied to the LUCIE emulator output. The deterministic baselines (bicubic, SFNO-SR, UNet-SR) smooth out the sharp orographic rainfall band along the Himalayan foothills and over northeastern India, whereas the diffusion-based models (Conditional EDM and DPS) recover the localized high-precipitation structure visible in ERA5. Color scale is shared across all panels.}
  \label{fig:MAM}
\end{figure}

To disentangle errors introduced by the downscaling step from those inherited from the LUCIE emulator, we additionally evaluate the diffusion models in a perfect-prognosis (PP) setting, in which they downscale coarsened ERA5 inputs rather than LUCIE outputs (bottom two rows of Figures~\ref{fig1:climatology} and~\ref{fig1b:climatology_bias}, labelled T30\_ERA5). In this configuration the input coarse state is, by construction, free of emulator bias and internal-variability mismatch with the ERA5 target. For the Conditional EDM the resulting climatological bias maps (Figure~\ref{fig1b:climatology_bias}) are substantially smaller and spatially smoother than those of the corresponding LUCIE-conditioned rows for all four variables, with residuals approaching zero across most of the globe and the largest remaining structure confined to faint precipitation features along the deep tropics. This indicates that the Conditional EDM itself introduces little climatological bias, and that the larger biases seen when downscaling LUCIE---for example, the high-latitude temperature and tropical wind structures---are predominantly inherited from the coarse LUCIE state rather than generated by the SR step. The conclusion does not extend to DPS, however. Even in this clean-input PP setting, DPS retains non-trivial residuals (Table~\ref{tab:pp_climatology_rmse}: 0.773~K for temperature, 0.969~m\,s$^{-1}$ for $u$-wind, and 0.597~m\,s$^{-1}$ for $v$-wind), well above the Conditional EDM's own PP residuals. For DPS, therefore, the super-resolution step itself injects appreciable bias even when the coarse input is free of emulator error; we trace this to its weak coarse-conditioning fidelity in Sections~\ref{sec:cond_vs_dps} and~\ref{sec:discussion}. The PP diagnostic thus addresses the concern that a downscaler might appear skillful for the wrong reasons: it confirms that the ERA5-trained Conditional EDM transfers to LUCIE inference without introducing significant spurious bias, while showing that DPS does not yet share this property.

Beyond global averages, the fidelity of the EDM models is particularly evident when examining seasonal temperature extremes over the Contiguous United States (CONUS) shown in Figures~\ref{fig:DJF} (DJF) and~\ref{fig:JJA} (JJA). In the June--July--August (JJA) temperature mean (Figure~\ref{fig:JJA}), the ERA5 ground truth reveals sharp thermal gradients driven by the complex topography of the Rocky Mountains and the Sierra Nevada. While the Bicubic, UNet-SR, and SFNO-SR approaches provide a generalized heat distribution, they suffer from spectral blurring and fail to resolve the cool-temperature ``islands" associated with high-altitude peaks. In contrast, the EDM approaches recover these fine-scale structures with high precision, accurately capturing the thermal contrast in the Intermountain West and the Central Valley of California, which matters for downstream applications such as wildfire-risk assessment where sub-grid temperature variations are decisive. During the December--January--February (DJF) period (Figure~\ref{fig:DJF}), the EDM models demonstrate superior skill in reconstructing cold-air pooling over the Northern Plains and the sharp transition zones along the Appalachian range. The same advantage holds for regional precipitation. Over the Indian subcontinent during the pre-monsoon (March--April--May, MAM) season (Figure~\ref{fig:MAM}), both diffusion-based downscalers recover the sharp orographic precipitation band along the Himalayan foothills and the localized maximum over northeastern India that are present in ERA5 but smoothed into a broad, diffuse pattern by the bicubic, UNet-SR, and SFNO-SR baselines.

\begin{figure}[h]
  \centering
  \includegraphics[width=\linewidth]{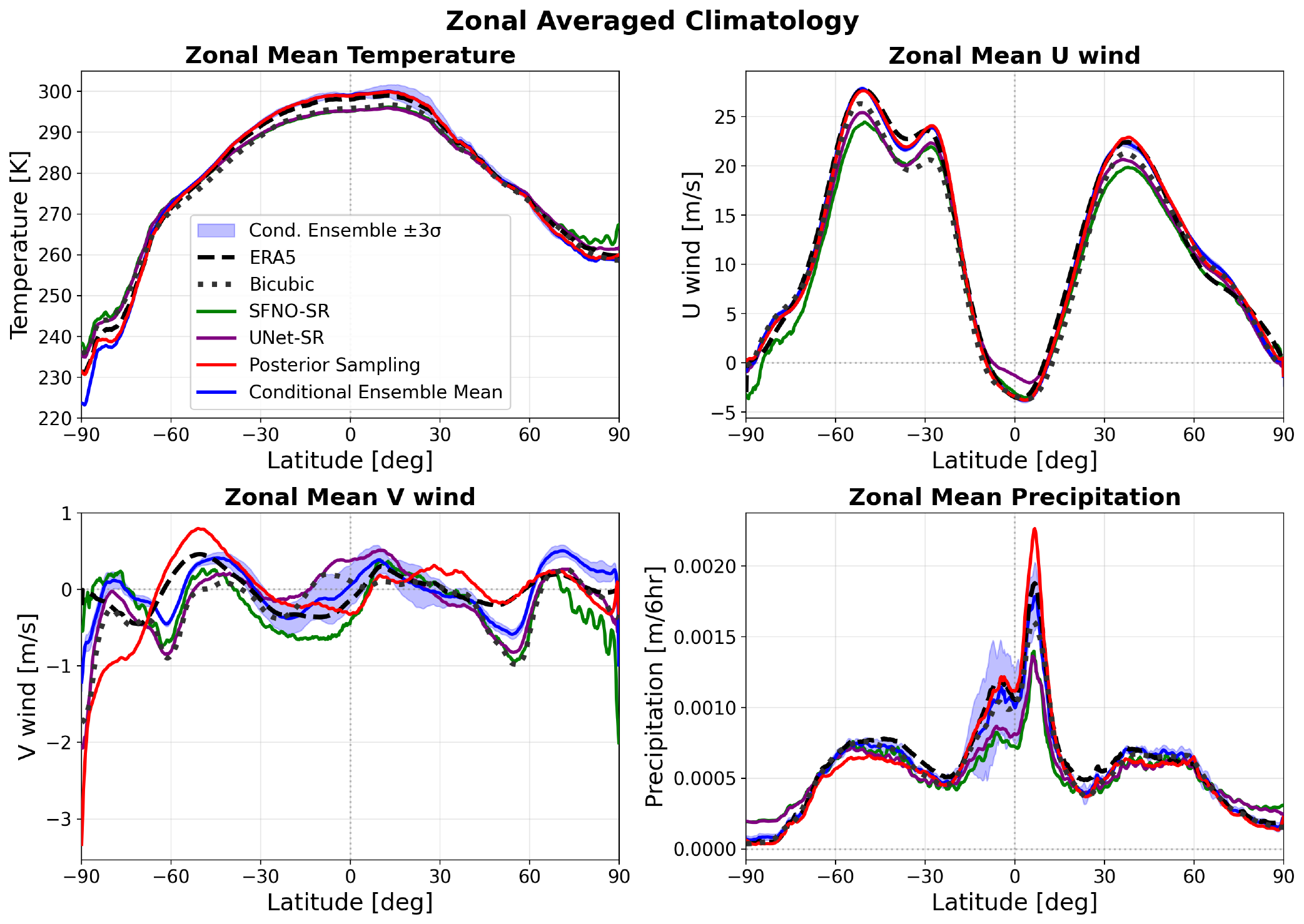}
  \caption{Latitude-weighted zonal climatology of temperature, $u$-wind, $v$-wind, and precipitation from 2010 to 2018. Each diffusion-based curve is the ensemble mean and the shaded band is the ensemble spread, in blue for the Conditional EDM and orange for DPS. The DPS spread is much narrower than the Conditional EDM spread; this is not a sign of higher confidence but of an overconfident, under-dispersed ensemble, as quantified in Section~\ref{sec:calibration} (Table~\ref{tab:ssr}). The meridional-wind bias in particular lies outside the Conditional EDM spread, and is shown by the re-coarsening check (Figure~\ref{fig:recoarsened_zonal}) to be inherited from the LUCIE $v$-wind input.}
  \label{fig:zonal}
\end{figure}

\begin{figure}[h!]
  \centering
  \includegraphics[width=\linewidth]{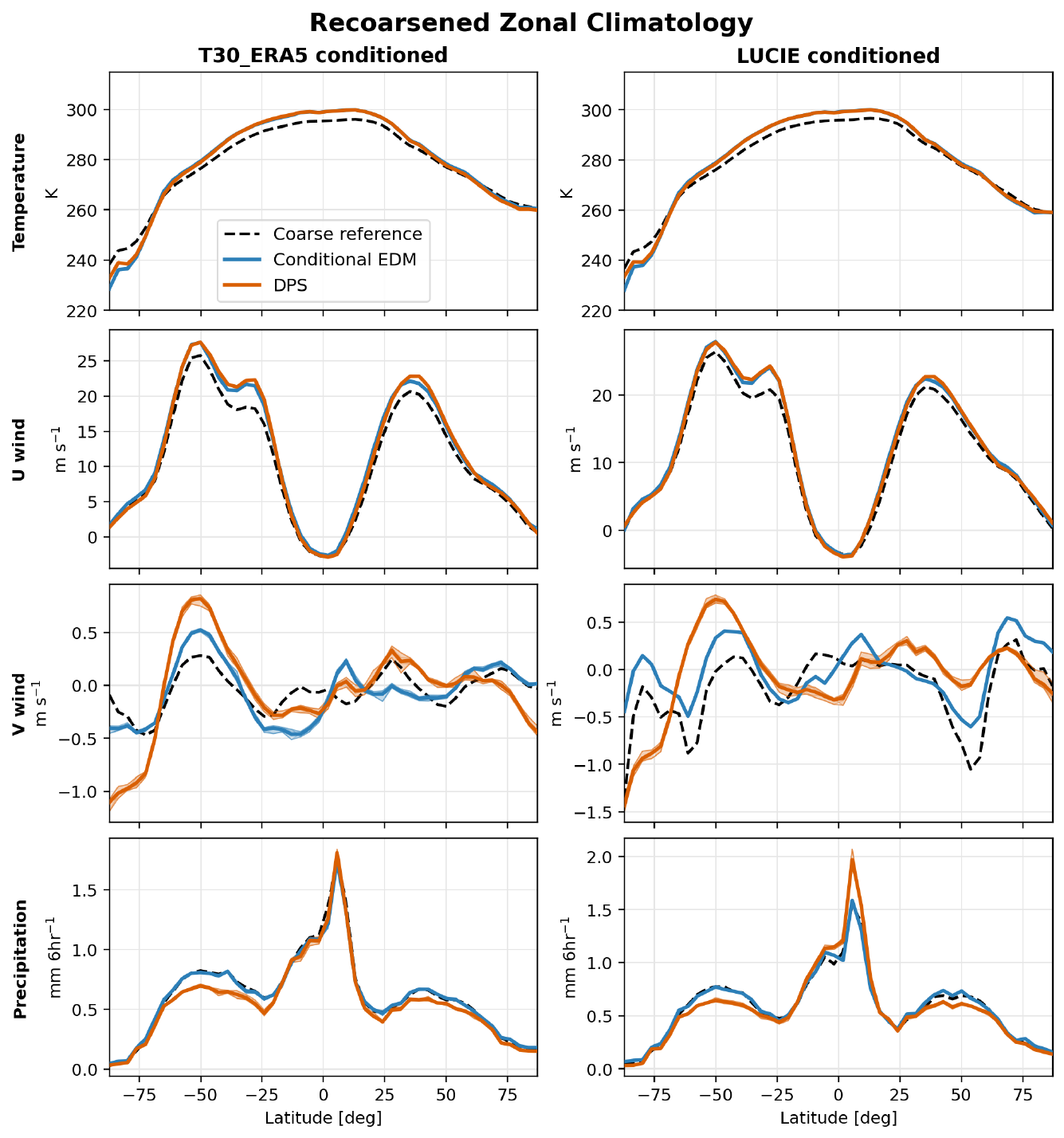}
  \caption{Re-coarsening consistency check for the zonal-mean climatology of temperature, $u$-wind, $v$-wind, and precipitation (2010--2018). Super-resolved fields are regridded back to the T30 coarse grid and compared against the coarse reference actually supplied to each model (dashed black). The reference differs by column: \textbf{(left)} T30-coarsened ERA5 in the perfect-prognosis setting, and \textbf{(right)} the native low-resolution LUCIE output in the imperfect-input setting. Blue and orange curves show the Conditional EDM and DPS ensemble means, respectively, with shaded bands indicating the ensemble spread. Close agreement with
the dashed reference indicates that the downscaler preserves the large-scale state it was
given. The Conditional EDM tracks the reference more closely than DPS---most visibly for
precipitation---consistent with its lower errors elsewhere in this study. The principal
exception is the meridional ($v$) wind in the LUCIE-conditioned column: because LUCIE's
$v$-wind is its weakest field and lies outside the distribution of coarse states seen during
ERA5 training, the ERA5-trained models reconstruct a $v$-wind that no longer re-coarsens to
the LUCIE input---an effect not seen in the perfect-prognosis column, where the same models
preserve a clean $v$-wind faithfully.}
  \label{fig:recoarsened_zonal}
\end{figure}

The zonal mean profiles in Figure~\ref{fig:zonal} offer an easier way to see how these models differ on a global scale. While the spatial maps shown earlier focus on regional details, these latitudinal graphs provide a clear quantitative look at how well each model follows the ERA5 ground truth. Bicubic interpolation, SFNO-SR, and UNet-SR follow the general patterns of the atmosphere, but they are clearly less accurate than the Conditional EDM, especially when it comes to capturing sharp changes in tropical rainfall and wind patterns. The most visible discrepancy is in the meridional wind, where the bias lies outside the Conditional EDM ensemble-spread band; we examine this further with a re-coarsening check below (Figure~\ref{fig:recoarsened_zonal}) and quantify the associated ensemble under-dispersion in Section~\ref{sec:calibration}.

In summary, the deterministic baselines retain substantial bias at the regional scale, whereas the generative downscalers recover much of the fine-scale structure and reduce this bias. They are not perfect, however: as the re-coarsening and calibration analyses below show, both diffusion models leave residual biases and produce under-dispersed ensembles.

\begin{table}[ht]
\centering
\caption{Latitude-weighted RMSE and Temporal Standard Deviation (STD) for the deterministic baselines and the generative diffusion downscalers. Values are shown as RMSE (STD). The lowest RMSE value for each variable is in bold. Deterministic methods are marked with $\dagger$ and generative (ensemble) methods with $\ast$.}
\begin{tabular}{l ccc cc}
\toprule
& \multicolumn{3}{c}{\textbf{Deterministic}~$^{\dagger}$} & \multicolumn{2}{c}{\textbf{Generative}~$^{\ast}$} \\
\cmidrule(lr){2-4} \cmidrule(lr){5-6}
Variable & Bicubic$^{\dagger}$ & SFNO-SR$^{\dagger}$ & UNet-SR$^{\dagger}$ & Conditional EDM$^{\ast}$ & DPS$^{\ast}$ \\
\midrule
Temperature [K]        & 2.581 (5.34)  & 2.497 (4.99)  & 2.427 (4.91)  & 1.357 (6.07)           & \textbf{1.237} (5.94) \\
$u$-wind [m s$^{-1}$]  & 2.358 (11.61) & 2.387 (11.79) & 2.000 (11.54) & 1.481 (12.97)          & \textbf{1.390} (11.83) \\
$v$-wind [m s$^{-1}$]  & 0.933 (10.48) & 0.780 (10.56) & 0.815 (10.20) & \textbf{0.770} (11.50) & 0.798 (10.76) \\
Precipitation**        & 0.249 (0.8)   & 0.325 (0.7)   & 0.301 (0.6)   & \textbf{0.212} (1.1)   & 0.221 (1.3) \\
\bottomrule
\end{tabular}
\label{tab:rmse}
\begin{flushleft}
\small $^{\dagger}$ Deterministic single-estimate methods. \quad
$^{\ast}$ Generative ensemble methods (values are ensemble-mean RMSE/STD over 16 members).\\
** Precipitation values for RMSE and STD are scaled by $10^3$ for readability ($10^{-3}$ m/6hr).
\end{flushleft}
\end{table}

To verify that the super-resolution step preserves the large-scale information provided by the coarse input---rather than altering it while adding fine-scale detail---we regrid the super-resolved fields back to the T30 coarse grid and compare the resulting zonal climatology against the coarse reference that was actually supplied to each model (Figure~\ref{fig:recoarsened_zonal}). The reference differs by column: in the perfect-prognosis case (left) it is the T30-coarsened ERA5 input, whereas in the imperfect-input case (right) it is the native low-resolution LUCIE output. An ideal downscaler should reproduce its own coarse reference upon recoarsening, since the fine-scale structure it introduces averages out at the coarse resolution. For temperature, zonal wind, and precipitation, both diffusion models recover the coarse reference well in both settings, confirming that the SR mapping is largely self-consistent with the low-resolution state it is given and does not grossly distort the large-scale climatology. For transparency, the agreement is not exact: visible residuals remain in both the perfect-prognosis and the LUCIE columns and for both models, indicating that neither framework reproduces its own coarse input perfectly even when that input is clean ERA5. The principal exception is the meridional $v$-wind. In the perfect-prognosis column the re-coarsened $v$-wind already shows a discernible departure from the coarsened-ERA5 reference, and in the LUCIE column both models depart substantially from the LUCIE $v$-wind they were conditioned on. The $v$-wind is the hardest field to reconstruct: it is LUCIE's weakest, lowest-amplitude field, with the smallest coarse-scale signal-to-noise ratio, so the ERA5-trained mapping has the least coarse-scale structure to anchor to, and the LUCIE $v$-wind in particular lies outside the distribution of coarse states seen during ERA5 training, so the model reconstructs a $v$-wind structure that no longer re-coarsens to the LUCIE input. The two generative frameworks also behave differently. The Conditional EDM, which is trained to match its coarse conditioning, tracks the reference more closely. DPS, by contrast, produces re-coarsened $v$-wind profiles that look similar in the perfect-prognosis and LUCIE columns despite the two coarse inputs being different, which indicates that DPS leans on its learned prior more than on the likelihood and therefore follows its coarse conditioning only weakly---it does not fully satisfy its own measurement operator. The same pattern appears for precipitation, where DPS reproduces the zonal-mean profile slightly less accurately than the Conditional EDM, particularly in the extratropics, consistent with the larger precipitation error reported for DPS elsewhere in this study. We return to these residual biases and the DPS measurement-operator issue in Section~\ref{sec:discussion}.

\subsection{Conditional EDM versus Diffusion Posterior Sampling}
\label{sec:cond_vs_dps}

\begin{figure}[ht!]
  \centering
  \includegraphics[width=\linewidth]{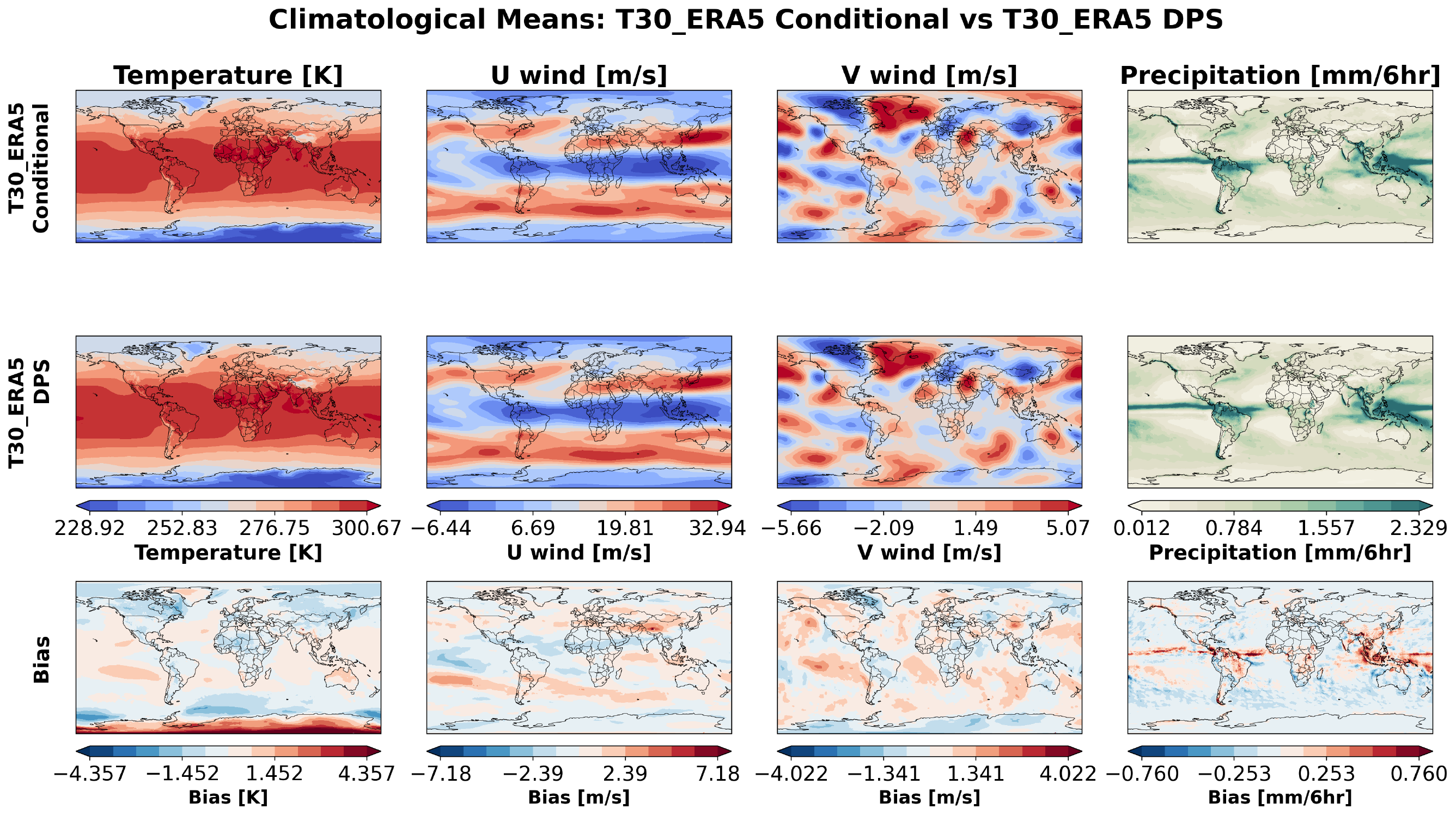}
  \caption{Perfect-prognosis (PP) comparison of climatological-mean fields produced by the Conditional EDM (top row) and Diffusion Posterior Sampling (DPS, middle row) when both models are applied to T30-coarsened ERA5 inputs (2010--2018). Columns correspond to temperature, $u$-wind, $v$-wind, and precipitation. The bottom row shows the difference (Conditional EDM $-$ DPS). Both models recover the large-scale structure; differences are concentrated at regional scales and over the deep tropics.}
  \label{fig:pp_cond_dps}
\end{figure}

\begin{figure}[ht!]
  \centering
  \includegraphics[width=\linewidth]{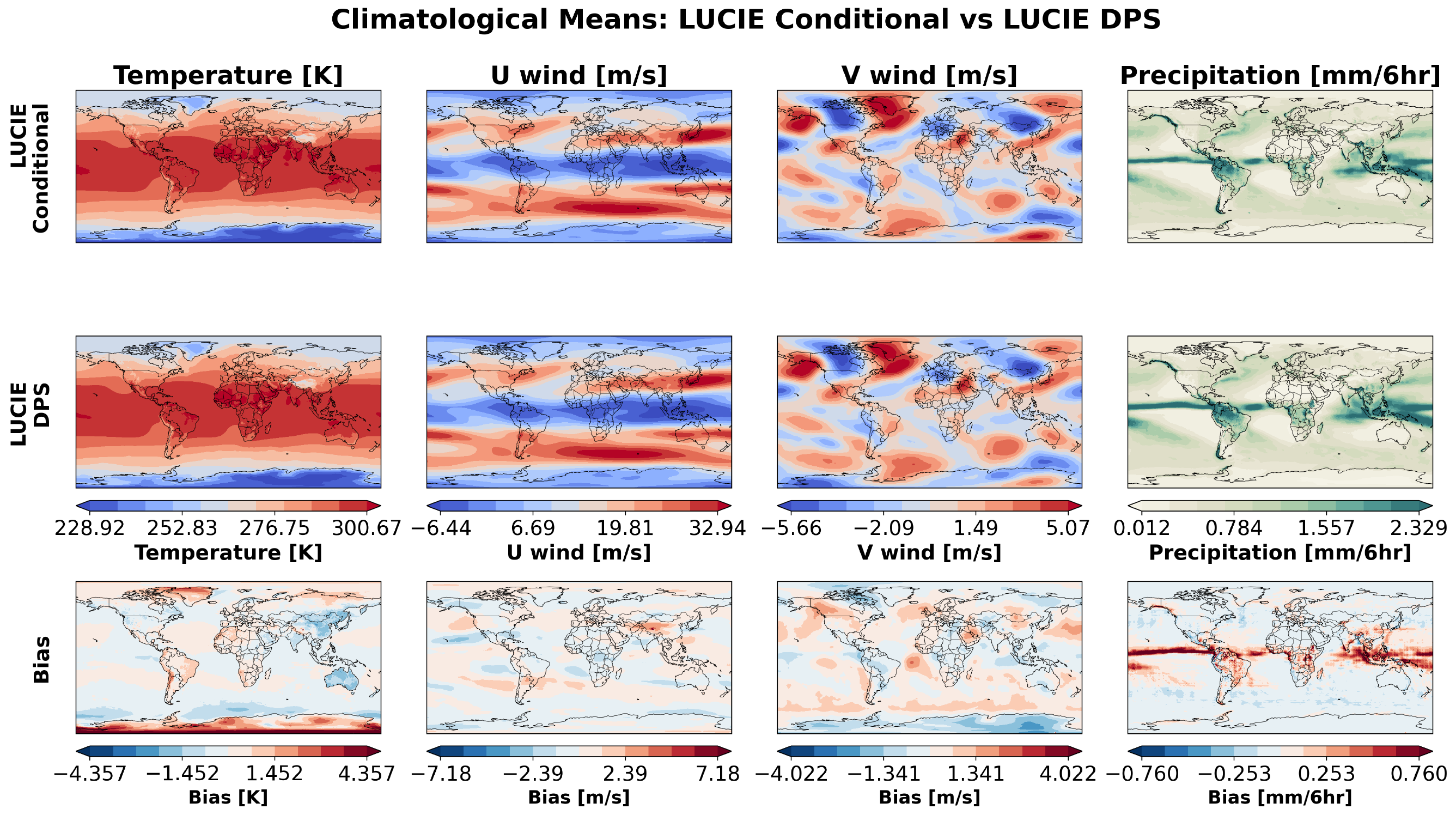}
  \caption{Imperfect-input comparison of climatological-mean fields produced by the Conditional EDM (top row) and Diffusion Posterior Sampling (DPS, middle row) when both models are applied to LUCIE outputs (2010--2018). Columns correspond to temperature, $u$-wind, $v$-wind, and precipitation. The bottom row shows the difference (Conditional EDM $-$ DPS). The two frameworks reproduce a similar large-scale climatology, with localized discrepancies over regions of sharp dynamical activity and the deep-tropical convergence zones.}
  \label{fig:lucie_cond_dps}
\end{figure}

Having established the climatological skill of the diffusion-based downscalers, we now compare the two generative frameworks directly: the Conditional EDM, which learns the high-resolution conditional distribution explicitly, and the Diffusion Posterior Sampling (DPS) approach, which guides an unconditional prior toward the coarse measurement at inference time. Because the posterior sampling formulation enforces consistency with the low-resolution field through an explicit likelihood term, it is natural to ask whether this constraint translates into improved climatological fidelity relative to the conditional model. We examine this question in both the imperfect-input setting, where the downscalers act on LUCIE outputs (Figure~\ref{fig:lucie_cond_dps}), and the perfect-prognosis (PP) setting, where they act on coarsened ERA5 inputs (Figure~\ref{fig:pp_cond_dps}).

In both settings the two models reproduce the same large-scale climatological structure---the meridional temperature gradient, the subtropical and mid-latitude jets in the zonal wind, and the ITCZ precipitation band are all recovered by both frameworks. The differences between them, shown in the bottom row of each figure, are largely confined to the regional scale. For temperature, the Conditional EDM and DPS agree to within a fraction of a degree across most of the globe, with the largest discrepancies appearing over the high-latitude Southern Ocean and the polar margins, where the absence of strong coarse-scale gradients leaves the reconstruction least constrained. The wind components show a similar pattern of broad agreement with localized differences over regions of sharp dynamical activity, while the precipitation difference is concentrated along the deep-tropical convergence zones, where both models must synthesize the most fine-scale variance. Notably, the inter-model difference is markedly larger in the perfect-prognosis setting (Figure~\ref{fig:pp_cond_dps}) than under LUCIE (Figure~\ref{fig:lucie_cond_dps}): the Conditional EDM has much lower bias than DPS in the perfect-prognosis case, whereas under LUCIE inputs the two converge and the gap shrinks. This mirrors the RMSE ordering reported below and is explained by the same mechanism.

Quantitatively, the PP experiment provides an upper-bound test of the downscaling models under the most favorable input distribution. Table~\ref{tab:pp_climatology_rmse} reports the latitude-weighted RMSE of the PP climatological means against ERA5. Both generative methods perform better in this setting than when applied to LUCIE outputs, as expected, because the coarse inputs are generated from ERA5 rather than from an imperfect emulator. This distinction is especially important for the Conditional EDM: in the PP case, the model is conditioned on the same T30 ERA5 fields used during training, so the experiment is close to giving the model its prior distribution. Under this favorable setup, the Conditional EDM achieves the lower error for every variable, with RMSEs of 0.260~K for temperature, 0.234~m\,s$^{-1}$ for $u$-wind, 0.176~m\,s$^{-1}$ for $v$-wind, and 0.089~mm/6hr for precipitation, compared with 0.773~K, 0.969~m\,s$^{-1}$, 0.597~m\,s$^{-1}$, and 0.158~mm/6hr for DPS.

The two generative frameworks differ in how they use the coarse input, and this directly explains the ordering of the errors. The Conditional EDM has no explicit separation of a prior and a likelihood: it learns the conditional distribution $p(\mathbf{x}\mid\mathbf{y})$ directly from paired T30~ERA5~$\rightarrow$~high-resolution~ERA5 samples. The perfect-prognosis setting is an idealized upper bound, in which the coarse input is, by construction, drawn from the same ERA5 distribution used in training and is free of emulator bias; there the direct conditional map is most effective and the Conditional EDM pulls well ahead of DPS, producing the large bias gap visible in the difference maps. Under LUCIE the coarse input instead carries the emulator's documented climatological biases, which both downscalers inherit, so absolute errors rise for both models and the inter-model gap narrows. For the Conditional EDM, the re-coarsening and bias-map analyses confirm that the SR step adds little bias of its own, so its residual LUCIE-side differences trace back to the coarse LUCIE state rather than to the downscaler. DPS, by contrast, learns only an unconditional high-resolution prior and applies a coarse-resolution likelihood at sampling time. The likelihood term is also easier to satisfy in PP than against LUCIE, so DPS improves under PP as well; the remaining gap behind the Conditional EDM in the PP setting reflects the additional sampling variability of enforcing the likelihood at inference rather than absorbing it into the trained conditioning path, together with DPS's tendency to follow its coarse conditioning only weakly (the re-coarsening check in Section~\ref{sec:climatology}). We return to the calibration consequences of this comparison in Section~\ref{sec:calibration} and to its broader implications in Section~\ref{sec:discussion}. The flexibility of DPS---and the reason we retain it in this study---is that the same trained prior can be combined with different coarse data sources or physically-informed measurement operators at inference time, enabling tasks such as observation-constrained downscaling without retraining \citep{chakraborty2026multimodal}. Such inference-time correction is not the focus of this paper, but it explains why DPS gives up accuracy relative to the dedicated conditional model in the perfect-prognosis setting and in conditioning fidelity: it is solving a strictly more general problem.

\begin{table}[ht]
\centering
\caption{Latitude-weighted RMSE of perfect-prognosis (T30\_ERA5-conditioned) climatological means against ERA5. Both methods are generative ensemble downscalers ($\ast$); the values reported are ensemble-mean RMSE.}
\label{tab:pp_climatology_rmse}
\begin{tabular}{lcc}
\toprule
Variable & Conditional EDM$^{\ast}$ & DPS$^{\ast}$ \\
\midrule
Temperature [K]         & \textbf{0.260} & 0.773 \\
$u$-wind [m s$^{-1}$]   & \textbf{0.234} & 0.969 \\
$v$-wind [m s$^{-1}$]   & \textbf{0.176} & 0.597 \\
Precipitation [mm/6hr]  & \textbf{0.089} & 0.158 \\
\bottomrule
\end{tabular}
\end{table}

Lower ensemble-mean RMSE does not by itself certify a better generative model, since a probabilistic downscaler should reproduce the \emph{distribution} of plausible states rather than minimize point-wise error \citep{harris2022generative, schillinger2025enscale}. The RMSE comparison should therefore be read alongside the spectral, distributional, and calibration diagnostics presented in Section~\ref{sec:spectrum}. With that caveat, the present comparison indicates that for the climatological-mean fields that are the focus of this study, the conditional formulation is, in the perfect-prognosis setting, both more accurate and, as discussed in Section~\ref{sec:cost}, substantially cheaper to sample, reinforcing its role as the more practical choice for high-resolution emulation at scale; DPS retains its distinct advantage of admitting observational constraints at inference time without retraining.

\subsection{Power spectrum and extreme events}
\label{sec:spectrum}
A significant strength of LUCIE is its ability to reproduce key statistical properties—including the zonal power spectrum, variance distribution, and return-period behavior—directly on the coarse T30 grid. Prior analyses have shown that LUCIE preserves the observed spectral slope across a wide range of scales and captures realistic distributional tails for temperature, winds, and precipitation. Therefore, a critical goal of the super-resolution in this work is not to correct deficiencies in LUCIE but to upscale its outputs while maintaining these desirable statistical properties on a finer grid.

\begin{figure}[!ht]
  \centering
  \includegraphics[width=\linewidth]{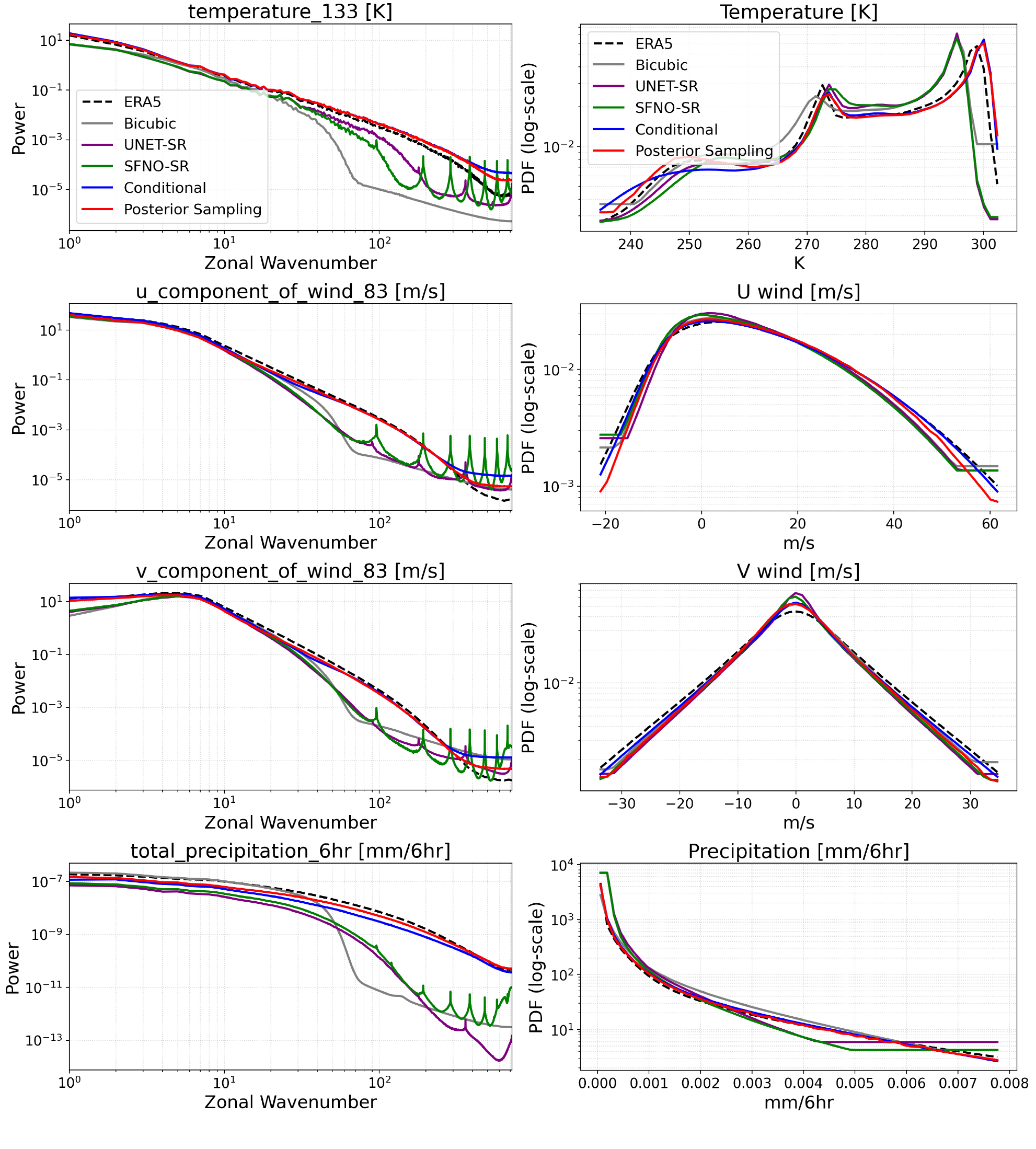}
  \caption{Comparison of mean spectral magnitudes across wavenumbers (left column) and probability density functions (right column) for near-surface temperature, zonal wind, meridional wind, and precipitation. Curves correspond to bicubic interpolation, SFNO-SR, UNet-SR, the Conditional EDM, and Diffusion Posterior Sampling (DPS). The Conditional EDM and DPS match the high-resolution ERA5 power spectra closely, with DPS tracking the reference more closely in the low-wavenumber band, while the deterministic baselines fail to recover the high-wavenumber tail.}
  \label{fig3:metrics}
\end{figure}

To evaluate this, we examine the zonal power spectra (left column of Figure~\ref{fig3:metrics}) and the corresponding PDFs on the right column. Bicubic interpolation, used as a baseline, exhibits severe damping of intermediate and high-wavenumber variance, reflecting the strong smoothing inherent to polynomial interpolation. In the low-wavenumber regime, SFNO-SR aligns reasonably well with the ERA5 reference. However, as the spatial scale decreases, its power spectrum begins to exhibit physically unrealistic wavy oscillations and periodic spikes in the high-wavenumber realm. These artifacts likely stem from the Fourier-based nature of the architecture, where aliasing or Gibbs-like phenomena can introduce artificial periodicities when reconstructing fine-scale turbulence from a truncated spectral basis \citep{bonev2023spherical}. In contrast, both EDM-based methods retain much more of the original spectral structure. The Conditional EDM restores a realistic spectral slope and captures a substantial portion of the missing high-wavenumber power, though it underestimates the precipitation power content at low wavenumbers ($k<100$) relative to ERA5. DPS tracks the reference spectra more closely in this low-wavenumber band, reproducing both the large-scale slope and the tail. We note that this spectral result and the re-coarsening result of Section~\ref{sec:climatology} measure different properties and are not in conflict. The power spectrum is a statistical, variance-based diagnostic: it asks whether the field carries the correct \emph{amount} of variability at each scale, averaged over the record. The re-coarsening check is a per-sample, mean-matching diagnostic: it asks whether each generated field, when coarsened, returns the \emph{specific} coarse state it was conditioned on. A sampler that leans on its learned prior, as DPS does, can produce fields with realistic large-scale variance---hence a good low-wavenumber spectrum---while still failing to reproduce the particular coarse input it was given, hence the weaker re-coarsening agreement and conditioning fidelity discussed in Section~\ref{sec:climatology}. The two findings therefore describe complementary aspects of DPS rather than contradicting each other. This indicates that diffusion-based super-resolution can preserve LUCIE’s long-term variability during downscaling rather than artificially smoothing or amplifying it. The right column of Figure~\ref{fig3:metrics} demonstrates the models' ability to capture the true distribution of atmospheric variables. The diffusion models capture the heavy tails and extremes much more accurately than bicubic interpolation. This is not surprising due to the ability of generative diffusion models to reconstruct fine details in the spatial field.

\subsection{EOF of zonal wind}

To evaluate how well each model captures dominant atmospheric patterns, an EOF analysis was performed on the zonal wind component. The first EOF mode (EOF1) represents the primary spatial pattern of variability, specifically the Northern Annular Mode (NAM) during the Northern Hemisphere winter (DJF) and the Southern Annular Mode (SAM) during the Southern Hemisphere winter (JJA).

\begin{figure}[!ht]
  \centering
  \includegraphics[width=\linewidth]{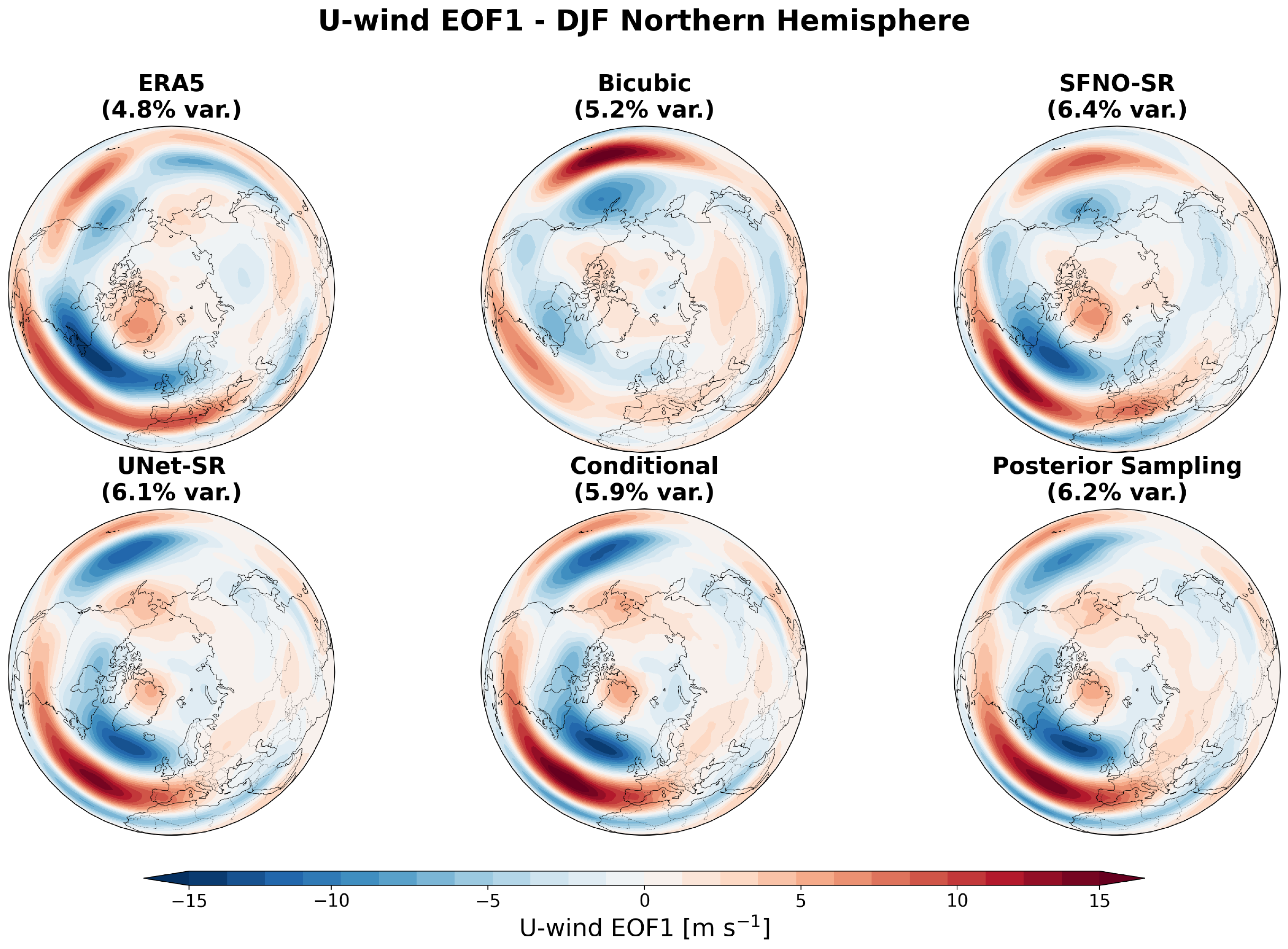}
  \caption{Comparison of the leading EOF of the DJF Northern Hemisphere zonal wind, corresponding to the Northern Annular Mode (NAM), for ERA5 and each downscaling configuration. Percentages indicate the variance explained by EOF1 in each panel; the ERA5 reference is 4.8\%. The EOF is computed over the Northern Hemisphere using a latitude weighting of $\sqrt{\cos(\text{latitude})}$ to account for meridional area differences.}
  \label{fig4:nam}
\end{figure}

In the Northern Hemisphere (Figure~\ref{fig4:nam}), ERA5 EOF1 shows a characteristic tri-polar structure, explaining 4.8\% of the total variance. While Bicubic interpolation overestimates the northernmost anomaly at 5.2\%, the deterministic SFNO-SR and UNet-SR models recover the spatial structure but overestimate variance at 6.4\% and 6.1\%, respectively. The generative models—Conditional EDM (5.9\%) and DPS (6.2\%)—reproduce the spatial pattern and centers of action most faithfully, although they overestimate the explained variance relative to ERA5 (Bicubic, at 5.2\%, is in fact closest to the 4.8\% reference value). The UNet-SR, Conditional EDM, and DPS models show internally consistent EOF patterns that do not perfectly match the ERA5 reference, which we interpret as the SR step reproducing the dynamics implied by its coarse input without introducing secondary distortions.

\begin{figure}[!ht]
  \centering
  \includegraphics[width=\linewidth]{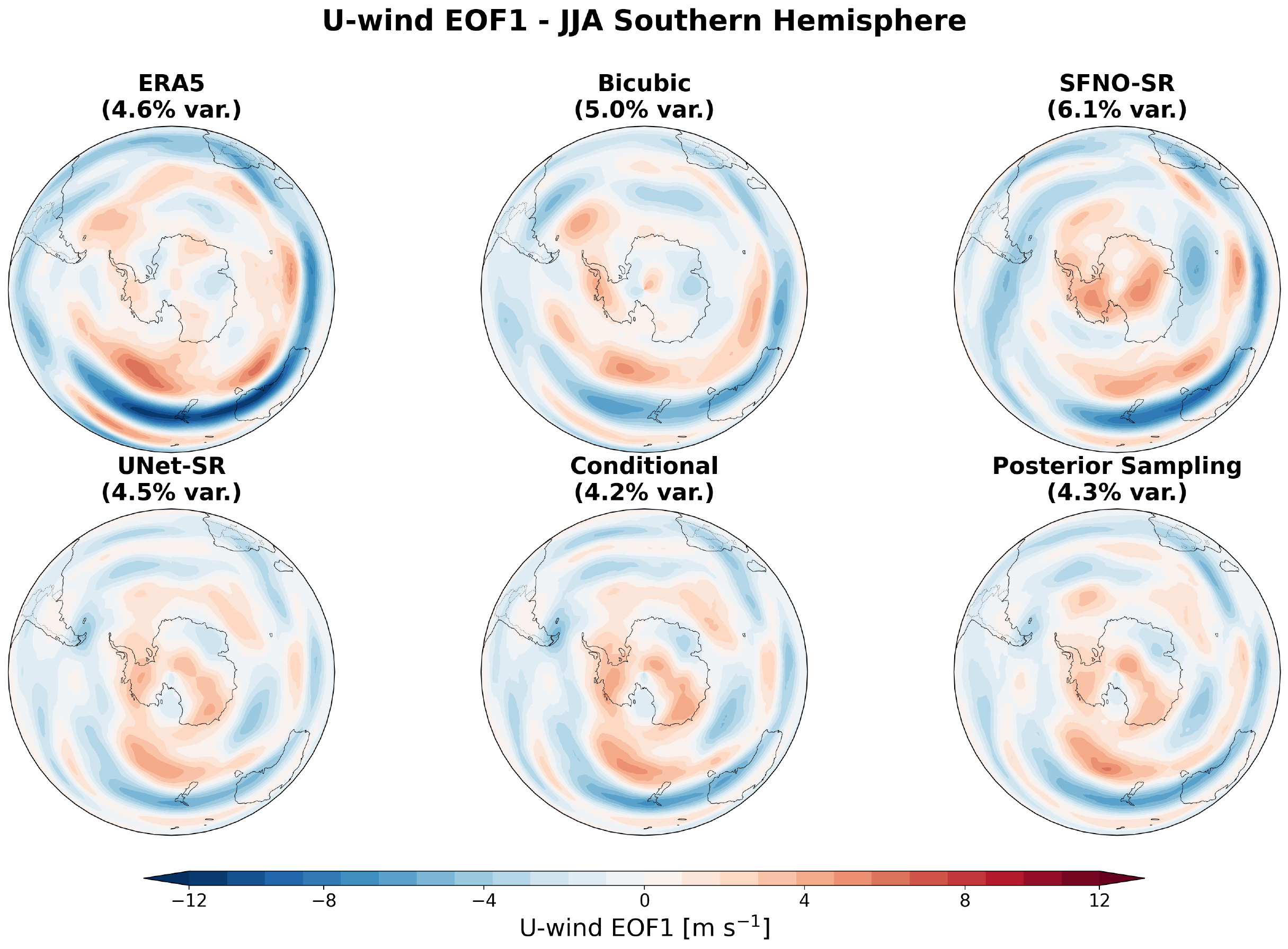}
  \caption{Comparison of the leading EOF of the JJA Southern Hemisphere zonal wind, corresponding to the Southern Annular Mode (SAM), for ERA5 and each downscaling configuration. Percentages indicate the variance explained by EOF1 in each panel; the ERA5 reference is 4.6\%. The EOF is computed over the Southern Hemisphere using a latitude weighting of $\sqrt{\cos(\text{latitude})}$ to account for meridional area differences.}
  \label{fig5:sam}
\end{figure}

For the Southern Hemisphere (Figure~\ref{fig5:sam}), EOF1 illustrates the Southern Annular Mode (SAM), explaining 4.6\% of the variance in the ERA5 data. All the models capture the general spatial pattern of ERA5. Both Bicubic and SFNO-SR models exhibit a spatial shift in the centers of action, with SFNO-SR notably overestimating variance at 6.1\%. The UNet-SR (4.5\%), Conditional EDM (4.2\%), and DPS (4.3\%) models accurately reconstruct the annular structure and the correct positioning of pressure centers. Similar to the Northern Hemisphere results, these models maintain a consistent internal EOF structure that appears to reflect the LUCIE dynamics, reconstructing the coarse-scale modes at high resolution without obvious added distortion. A matched, EOF-level decomposition of the LUCIE coarse fields against ERA5 is left for future work; given the climatological-bias and re-coarsening evidence in Section~\ref{sec:climatology}, we expect---but have not yet verified at the EOF level---that the residual EOF discrepancies under LUCIE conditioning are predominantly inherited from the LUCIE input rather than introduced by the SR step. We therefore present this attribution as a conjecture to be tested, not as an established result.

\subsection{Ensemble Calibration}
\label{sec:calibration}

The temporal standard deviations reported in Section~\ref{sec:climatology} show that
the diffusion models recover more climatological variability than the deterministic
baselines, but a larger spread is only meaningful if it matches the actual error of the
ensemble. To assess this, we compute the continuous ranked probability score (CRPS,
Table~\ref{tab:crps}) and the spread-skill ratio (SSR, Table~\ref{tab:ssr}) in both the
perfect-prognosis setting (T30\_ERA5 inputs evaluated against ERA5) and the
imperfect-input setting (LUCIE inputs). A well-calibrated ensemble has an SSR near unity;
values below one indicate under-dispersion.

\begin{table}[htbp]
\centering
\caption{Continuous Ranked Probability Score (CRPS) for the generative ($\ast$) Conditional EDM and
Diffusion Posterior Sampling (DPS) models, in the perfect-prognosis setting (T30\_ERA5
inputs, evaluated against ERA5) and the imperfect-input setting (LUCIE inputs). Lower is
better; the best (lowest) value for each variable within each setting is in bold. We report
CRPS only for the generative models, which are designed to produce a calibrated ensemble; the
deterministic baselines are single-point estimates and are assessed by RMSE in
Table~\ref{tab:rmse}.}
\label{tab:crps}
\begin{tabular}{llccccc}
\toprule
Method & Setting & Temperature [K] & $u$-wind [m s$^{-1}$] & $v$-wind [m s$^{-1}$] & TP [mm/6hr] \\
\midrule
Conditional EDM$^{\ast}$ & T30\_ERA5 & \textbf{0.565} & \textbf{1.651} & \textbf{1.554} & \textbf{0.351} \\
DPS$^{\ast}$             & T30\_ERA5 & 1.098 & 3.349 & 3.141 & 0.480 \\
Conditional EDM$^{\ast}$ & LUCIE     & 1.086 & \textbf{1.057} & \textbf{0.472} & \textbf{0.099} \\
DPS$^{\ast}$             & LUCIE     & \textbf{0.964} & 1.146 & 0.766 & 0.120 \\
\bottomrule
\end{tabular}
\end{table}

The SSR lies well below one for nearly all variables and both settings. In the
perfect-prognosis setting it ranges from roughly 0.55 to 0.70 for temperature and the
wind components and is lowest for precipitation (0.239 for the Conditional EDM, 0.302 for
DPS). Under LUCIE inputs the wind components are more under-dispersed still, with the
$u$-wind SSR falling to 0.229 for both models and DPS temperature dropping to 0.108. We
therefore conclude plainly that, as configured, neither diffusion model is well
calibrated: both are under-dispersed, and their ensemble spread systematically
underestimates the true error. The diffusion models add realistic fine-scale variance and
achieve low ensemble-mean error---CRPS in the perfect-prognosis setting favors the
Conditional EDM across all variables---but they do not produce reliable uncertainty
estimates in their current form.

\begin{table}[htbp]
\centering
\caption{Spread-skill ratio (SSR) for the generative ($\ast$) Conditional EDM and DPS models in the
perfect-prognosis (T30\_ERA5) and imperfect-input (LUCIE) settings. Values near unity
indicate calibration; below unity indicates under-dispersion (overconfidence). Because SSR is
not simply ``lower is better,'' the value closest to unity for each variable within each
setting is in bold. We report SSR only for the generative models, which are designed to
produce a calibrated ensemble; the deterministic baselines are single-point estimates and are
assessed by RMSE in Table~\ref{tab:rmse}.}
\label{tab:ssr}
\begin{tabular}{llccccc}
\toprule
Method & Setting & Temperature [K] & $u$-wind [m s$^{-1}$] & $v$-wind [m s$^{-1}$] & TP [mm/6hr] \\
\midrule
Conditional EDM$^{\ast}$ & T30\_ERA5 & 0.556 & 0.648 & \textbf{0.655} & 0.239 \\
DPS$^{\ast}$             & T30\_ERA5 & \textbf{0.599} & \textbf{0.703} & 0.638 & \textbf{0.302} \\
Conditional EDM$^{\ast}$ & LUCIE     & \textbf{0.700} & \textbf{0.229} & \textbf{0.473} & \textbf{0.583} \\
DPS$^{\ast}$             & LUCIE     & 0.108 & \textbf{0.229} & 0.274 & 0.418 \\
\bottomrule
\end{tabular}
\end{table}

\section{Discussion and Conclusions}\label{sec:discussion}
In this study, we evaluate the viability of a two-stage modeling framework: using super-resolution (SR) models to upscale the coarse-grid outputs of a lightweight climate emulator. While our climate emulator, LUCIE, excels at capturing large-scale dynamics and global statistics, its utility for regional-scale analysis is inherently limited by its T30 Gaussian grid. We demonstrate that the introduction of an SR model effectively bridges this gap, reconstructing fine-scale spatial structures such as temperature gradients in the Rocky Mountains that are implied by the large-scale state but unresolved on the coarse grid.

Super-resolution acts as a faithful spatial extension of the underlying emulator: the SR process does not merely smooth the coarse fields but restores the fine-scale statistics implied by the large-scale state. Deterministic baselines such as bicubic interpolation and SFNO-SR recover the general atmospheric patterns but lack the regional precision needed for impact studies. The diffusion-based models (Conditional EDM and DPS) recover localized regional features such as the sharp pre-monsoon orographic precipitation band over the Indian peninsula (Figure~\ref{fig:MAM}) and the seasonal temperature contrasts across the Intermountain West and the Sierra Nevada (Figures~\ref{fig:DJF}--\ref{fig:JJA}). These are the kinds of fine-scale features relevant to applications such as monsoon-dependent agriculture and wildfire-risk assessment, so the LUCIE-SR framework is a step toward bringing such applications within the reach of coarse-resolution emulator pipelines. We stress that this is a demonstration on stationary runs with the limitations noted below (imperfect calibration, residual bias, and the absence of a forced emulator), not yet an operational impact-assessment tool.

The atmospheric downscaling task is fundamentally one-to-many: a single coarse-scale state is consistent with many physically plausible high-resolution realizations, which makes a probabilistic formulation a natural fit. The ability of diffusion models to draw an ensemble of realizations provides a measure of sub-grid uncertainty that the deterministic baselines cannot offer. SFNO-SR is included in this comparison not because it is the strongest baseline---it is not, and it introduces high-wavenumber spectral artifacts (Figure~\ref{fig3:metrics})---but because it shares its operator backbone with the LUCIE emulator. That shared backbone is what makes SFNO-SR attractive as a stepping stone toward a future unified, end-to-end SFNO-based emulation-and-downscaling pipeline, even though in the present, non-unified setting the generative downscalers clearly outperform it.

Several limitations of the generative downscalers deserve emphasis, since they bound how the framework should be used and point to concrete next steps. First, neither diffusion ensemble is well calibrated. The spread-skill ratios in Table~\ref{tab:ssr} lie well below unity for nearly all variables and both settings, so the ensembles are under-dispersed and their spread systematically underestimates the true error. The framework therefore recovers fine-scale structure and low ensemble-mean error, but it does not yet deliver reliable uncertainty quantification; correcting this---through explicit spread calibration or tuning of the DPS prior-likelihood weighting---is an important direction for future work. Second, DPS does not fully satisfy its own measurement operator. The re-coarsening check (Figure~\ref{fig:recoarsened_zonal}) shows that DPS retains large-scale coarse biases and reconstructs coarse states that look similar whether it is given ERA5 or LUCIE inputs, indicating that it follows its likelihood only weakly and leans on its learned prior. This weak conditioning fidelity is the main reason the dedicated Conditional EDM, which absorbs the coarse conditioning into training, is the more practical and accurate choice for the present downscaling task: it attains lower error for all four variables in the perfect-prognosis RMSE comparison (Table~\ref{tab:pp_climatology_rmse}), tracks its coarse reference more faithfully in the re-coarsening check (Figure~\ref{fig:recoarsened_zonal}), and is substantially cheaper to sample (Section~\ref{sec:cost}). We note, however, that this advantage is not uniform across every metric: on raw ensemble-mean RMSE under LUCIE inputs (Table~\ref{tab:rmse}), DPS attains lower error for near-surface temperature and zonal wind, while the Conditional EDM is lower for the meridional wind and precipitation. The contrast suggests that a more accurate measurement operator and stronger likelihood weighting are needed before DPS's flexibility can be exploited without a loss of conditioning fidelity. Third, biases remain even in the most favorable perfect-prognosis setting, where the coarse input is clean ERA5 rather than an imperfect emulator: the re-coarsening and climatological-bias diagnostics still show non-trivial residuals, most prominently in the meridional wind and along the deep-tropical precipitation band. The super-resolution step is therefore not bias-free, and the $v$-wind in particular---LUCIE's weakest, lowest-amplitude field, which lies outside the coarse-state distribution seen during ERA5 training---remains the hardest field to reconstruct and a clear target for improvement.

We emphasize that machine learning climate emulations are inherently imperfect, owing to the chaotic nature of climate dynamics and unavoidable spectral biases. We use generative modeling to sample plausible high-resolution states and to diagnose how the emulator's coarse-scale biases propagate through the SR pipeline, not because generative models necessarily correct reanalysis-emulator discrepancies. The perfect-prognosis and re-coarsening diagnostics above support this reading: the larger biases under LUCIE conditioning are predominantly inherited from the coarse LUCIE state rather than introduced by the downscaler.

The computational costs of these models vary with their architecture (precise runtimes are given in Section~\ref{sec:cost}). The deterministic UNet-SR is the fastest option but produces visually smooth outputs that lack the fine-scale variability of high-resolution weather data. The diffusion models recover these details at higher inference cost. The Conditional EDM is the practical compromise: its inference cost is only marginally above the deterministic baseline, making it well suited to the large ensembles that LUCIE is designed for, and its training distribution matches its deployment distribution while sidestepping the need to construct a likelihood operator at sampling time. DPS is substantially more expensive but more flexible: because its prior is unconditional, the same trained model can be reused with a different coarse data source or measurement operator at inference time, enabling physically-informed corrections such as observation-constrained downscaling without retraining \citep{chakraborty2026multimodal}. We do not exercise that capability here, but it is what makes DPS a useful general-purpose prior alongside the conditional model.

Overall, our results support a modular modeling strategy: by decoupling the simulation of large-scale dynamics from the reconstruction of spatial details, we can use the computational efficiency of models like LUCIE without giving up regional precision. The downscaler is not intended to correct the climate physics produced by the emulator; it adds the spatial detail that the coarse emulator cannot resolve. This two-stage approach offers a scalable pathway for generating the high-resolution, ensemble-based data required for climate impact assessment and risk management. Because the downscaler is a per-timestep spatial post-processing step, it is in principle agnostic to whether its coarse input represents present-day or future climate, and would lift a forced future LUCIE trajectory to high resolution in the same way it does the stationary runs evaluated here; realizing this for actual climate projection therefore depends on extending the underlying emulator to support a forced, nonstationary response \citep{guan2025lucie}, which we leave for future work.

\section*{Open Research Section}

The codes used for training and inference are permanently archived on Zenodo:\\(https://zenodo.org/records/18627189)~\citep{guan_2026_18627189}. The T30 Gaussian gridded ERA5 dataset required for training and the 10-year LUCIE emulation required for inference are included in the zenodo link. The high resolution ERA5 is available at:\\
https://doi.org/10.5065/XV5R-5344 \citep{ncar_gdex_dataset_d633006}. 

%
\bibliography{references}

@article{SchwingshacklEtAl2024,
  author = {Schwingshackl, Clemens and Daloz, Anne Sophie and Iles, Carley and Aunan, Kristin and Sillmann, Jana},
  title = {High‐resolution projections of ambient heat for major European cities using different heat metrics},
  journal = {Natural Hazards and Earth System Sciences},
  year = {2024},
  volume = {24},
  pages = {331--354},
  doi = {10.5194/nhess-24-331-2024}
}

@article{TuEtAl2025_MODS,
  author = {Tu, Siwei and Xu, Jingyi and Yang, Weidong and Bai, Lei and Fei, Ben},
  title = {MODS: Multi‐source Observations Conditional Diffusion Model for Meteorological State Downscaling},
  journal = {arXiv preprint},
  year = {2025},
  eprint = {2506.14798},
  url={https://arxiv.org/abs/2506.14798}
}

@article{TuEtAl2025_SGD,
  author = {Tu, Siwei and Fei, Ben and Yang, Weidong and Ling, Fenghua and Chen, Hao and Liu, Zili and others},
  title = {Satellite Observations Guided Diffusion Model for Accurate Meteorological States at Arbitrary Resolution},
  journal = {arXiv preprint},
  year = {2025},
  eprint = {2502.07814},
  url = {https://arxiv.org/abs/2502.07814}
}

@article{HarderEtAl2025_RainShift,
  author = {Harder, Paula and Schmidt, Luca and Pelletier, Francis and Ludwig, Nicole and Chantry, Matthew and Lessig, Christian and Hernandez‐Garcia, Alex and Rolnick, David},
  title = {RainShift: A Benchmark for Precipitation Downscaling Across Geographies},
  journal = {arXiv preprint},
  year = {2025},
  eprint = {2507.04930},
  url={https://arxiv.org/abs/2507.04930}
}

@article{ReddyEtAl2025_LimitationSR,
  author = {Reddy, P. Jyoteeshkumar and Matear, Richard and Taylor, John and Thatcher, Marcus and others},
  title = {Limitation of super-resolution machine learning approach to precipitation downscaling},
  journal = {Scientific Reports},
  year = {2025},
  volume = {15},
  article = {30070},
  doi = {10.1038/s41598-025-05880-7}
}

@article{SkamarockEtAl2019,
  author = {Skamarock, W. C. and Klemp, J. B. and Dudhia, J. and Gill, D. O. and Barker, D. M. and Wang, W. and Powers, J. G.},
  title = {A Description of the Advanced Research WRF Model Version 4},
  journal = {NCAR Technical Note},
  year = {2019}
}

@article{Giorgi2019,
  author = {Giorgi, Filippo},
  title = {Thirty years of regional climate modeling: Where are we and where are we going after CORDEX?},
  journal = {Earth's Future},
  year = {2019},
  doi = {10.1029/2019EF001341}
}

@article{mardani2025residual,
  title={Residual corrective diffusion modeling for km-scale atmospheric downscaling},
  author={Mardani, Morteza and Brenowitz, Noah and Cohen, Yair and Pathak, Jaideep and Chen, Chieh-Yu and Liu, Cheng-Chin and Vahdat, Arash and Nabian, Mohammad Amin and Ge, Tao and Subramaniam, Akshay and others},
  journal={Communications Earth \& Environment},
  volume={6},
  number={1},
  pages={124},
  year={2025},
  publisher={Nature Publishing Group UK London}
}

@article{ho2020denoising,
  title={Denoising diffusion probabilistic models},
  author={Ho, Jonathan and Jain, Ajay and Abbeel, Pieter},
  journal={Advances in neural information processing systems},
  volume={33},
  pages={6840--6851},
  year={2020}
}

@article{song2020score,
  title={Score-based generative modeling through stochastic differential equations},
  author={Song, Yang and Sohl-Dickstein, Jascha and Kingma, Diederik P and Kumar, Abhishek and Ermon, Stefano and Poole, Ben},
  journal={arXiv preprint arXiv:2011.13456},
  year={2020}
}

@article{karras2022elucidating,
  title={Elucidating the design space of diffusion-based generative models},
  author={Karras, Tero and Aittala, Miika and Aila, Timo and Laine, Samuli},
  journal={Advances in neural information processing systems},
  volume={35},
  pages={26565--26577},
  year={2022}
}

@article{chung2022diffusion,
  title={Diffusion posterior sampling for general noisy inverse problems},
  author={Chung, Hyungjin and Kim, Jeongsol and Mccann, Michael T and Klasky, Marc L and Ye, Jong Chul},
  journal={arXiv preprint arXiv:2209.14687},
  year={2022}
}

@article{rozet2023score,
  title={Score-based data assimilation},
  author={Rozet, Fran{\c{c}}ois and Louppe, Gilles},
  journal={Advances in Neural Information Processing Systems},
  volume={36},
  pages={40521--40541},
  year={2023}
}

@article{brenowitz2025climate,
  title={Climate in a Bottle: Towards a Generative Foundation Model for the Kilometer-Scale Global Atmosphere},
  author={Brenowitz, Noah D and Ge, Tao and Subramaniam, Akshay and Gupta, Aayush and Hall, David M and Mardani, Morteza and Vahdat, Arash and Kashinath, Karthik and Pritchard, Michael S},
  journal={arXiv preprint arXiv:2505.06474},
  year={2025}
}

@inproceedings{sohl2015deep,
  title={Deep unsupervised learning using nonequilibrium thermodynamics},
  author={Sohl-Dickstein, Jascha and Weiss, Eric and Maheswaranathan, Niru and Ganguli, Surya},
  booktitle={International conference on machine learning},
  pages={2256--2265},
  year={2015},
  organization={pmlr}
}

@article{guan2025lucie,
  title={LUCIE: A lightweight uncoupled climate emulator with long-term stability and physical consistency},
  author={Guan, Haiwen and Arcomano, Troy and Chattopadhyay, Ashesh and Maulik, Romit},
  journal={Journal of Advances in Modeling Earth Systems},
  volume={17},
  number={11},
  pages={e2025MS005152},
  year={2025},
  publisher={Wiley Online Library}
}

@article{mardani2024residual,
  title={Residual diffusion modeling for km-scale atmospheric downscaling},
  author={Mardani, Morteza and Brenowitz, Noah and Cohen, Yair and Pathak, Jaideep and Chen, Chieh-Yu and Liu, Cheng-Chin and Vahdat, Arash and Kashinath, Karthik and Kautz, Jan and Pritchard, Mike},
  year={2024}
}

@article{stengel2020adversarial,
  title={Adversarial super-resolution of climatological wind and solar data},
  author={Stengel, Karen and Glaws, Andrew and Hettinger, Dylan and King, Ryan N},
  journal={Proceedings of the National Academy of Sciences},
  volume={117},
  number={29},
  pages={16805--16815},
  year={2020},
  publisher={National Academy of Sciences}
}

@article{lopez2025dynamical,
  title={Dynamical-generative downscaling of climate model ensembles},
  author={Lopez-Gomez, Ignacio and Wan, Zhong Yi and Zepeda-N{\'u}{\~n}ez, Leonardo and Schneider, Tapio and Anderson, John and Sha, Fei},
  journal={Proceedings of the National Academy of Sciences},
  volume={122},
  number={17},
  pages={e2420288122},
  year={2025},
  publisher={National Academy of Sciences}
}

@inproceedings{wei2023super,
  title={Super-resolution neural operator},
  author={Wei, Min and Zhang, Xuesong},
  booktitle={Proceedings of the IEEE/CVF Conference on Computer Vision and Pattern Recognition},
  pages={18247--18256},
  year={2023}
}

@inproceedings{vandal2017deepsd,
  title={Deepsd: Generating high resolution climate change projections through single image super-resolution},
  author={Vandal, Thomas and Kodra, Evan and Ganguly, Sangram and Michaelis, Andrew and Nemani, Ramakrishna and Ganguly, Auroop R},
  booktitle={Proceedings of the 23rd acm sigkdd international conference on knowledge discovery and data mining},
  pages={1663--1672},
  year={2017}
}

@inproceedings{bonev2023spherical,
  title={Spherical fourier neural operators: Learning stable dynamics on the sphere},
  author={Bonev, Boris and Kurth, Thorsten and Hundt, Christian and Pathak, Jaideep and Baust, Maximilian and Kashinath, Karthik and Anandkumar, Anima},
  booktitle={International conference on machine learning},
  pages={2806--2823},
  year={2023},
  organization={PMLR}
}

@article{hersbach2020era5,
  title={The ERA5 global reanalysis},
  author={Hersbach, Hans and Bell, Bill and Berrisford, Paul and Hirahara, Shoji and Hor{\'a}nyi, Andr{\'a}s and Mu{\~n}oz-Sabater, Joaqu{\'\i}n and Nicolas, Julien and Peubey, Carole and Radu, Raluca and Schepers, Dinand and others},
  journal={Quarterly journal of the royal meteorological society},
  volume={146},
  number={730},
  pages={1999--2049},
  year={2020},
  publisher={Wiley Online Library}
}

@article{arcomano2022hybrid,
  title={A hybrid approach to atmospheric modeling that combines machine learning with a physics-based numerical model},
  author={Arcomano, Troy and Szunyogh, Istvan and Wikner, Alexander and Pathak, Jaideep and Hunt, Brian R and Ott, Edward},
  journal={Journal of Advances in Modeling Earth Systems},
  volume={14},
  number={3},
  pages={e2021MS002712},
  year={2022},
  publisher={Wiley Online Library}
}

@inproceedings{ronneberger2015u,
  title={U-net: Convolutional networks for biomedical image segmentation},
  author={Ronneberger, Olaf and Fischer, Philipp and Brox, Thomas},
  booktitle={International Conference on Medical image computing and computer-assisted intervention},
  pages={234--241},
  year={2015},
  organization={Springer}
}

@article{sharma2022resdeepd,
  title={ResDeepD: A residual super-resolution network for deep downscaling of daily precipitation over India},
  author={Sharma, Sumanta Chandra Mishra and Mitra, Adway},
  journal={Environmental Data Science},
  volume={1},
  pages={e19},
  year={2022},
  publisher={Cambridge University Press}
}

@article{zhang2024super,
  title={Super Resolution On Global Weather Forecasts},
  author={Zhang, Lawrence and Yang, Adam and Amor, Rodz Andrie and Zhang, Bryan and Rao, Dhruv},
  journal={arXiv preprint arXiv:2409.11502},
  year={2024}
}

@article{chakraborty2026multimodal,
  title={Multimodal atmospheric super-resolution with deep generative models},
  author={Chakraborty, Dibyajyoti and Guan, Haiwen and Stock, Jason and Arcomano, Troy and Cervone, Guido and Maulik, Romit},
  journal={Machine Learning: Earth},
  volume={2},
  number={1},
  pages={015001},
  year={2026},
  publisher={IOP Publishing}
}

@book{maraun2018statistical,
  title={Statistical downscaling and bias correction for climate research},
  author={Maraun, Douglas and Widmann, Martin},
  year={2018},
  publisher={Cambridge University Press}
}

@article{rampal2024enhancing,
  title={Enhancing regional climate downscaling through advances in machine learning},
  author={Rampal, Neelesh and Hobeichi, Sanaa and Gibson, Peter B and Ba{\~n}o-Medina, Jorge and Abramowitz, Gab and Beucler, Tom and Gonz{\'a}lez-Abad, Jose and Chapman, William and Harder, Paula and Guti{\'e}rrez, Jos{\'e} Manuel},
  journal={Artificial Intelligence for the Earth Systems},
  volume={3},
  number={2},
  pages={230066},
  year={2024},
  publisher={American Meteorological Society}
}

@article{bano2020configuration,
  title={Configuration and intercomparison of deep learning neural models for statistical downscaling},
  author={Ba{\~n}o-Medina, Jorge and Manzanas, Rodrigo and Guti{\'e}rrez, Jos{\'e} Manuel},
  journal={Geoscientific Model Development},
  volume={13},
  number={4},
  pages={2109--2124},
  year={2020},
  publisher={Copernicus GmbH}
}

@article{zhu2020gan,
  title={GAN-based image super-resolution with a novel quality loss},
  author={Zhu, Xining and Zhang, Lin and Zhang, Lijun and Liu, Xiao and Shen, Ying and Zhao, Shengjie},
  journal={Mathematical Problems in Engineering},
  volume={2020},
  number={1},
  pages={5217429},
  year={2020},
  publisher={Wiley Online Library}
}

@article{jiang2023efficient,
  title={Efficient super-resolution of near-surface climate modeling using the Fourier neural operator},
  author={Jiang, Peishi and Yang, Zhao and Wang, Jiali and Huang, Chenfu and Xue, Pengfei and Chakraborty, TC and Chen, Xingyuan and Qian, Yun},
  journal={Journal of Advances in Modeling Earth Systems},
  volume={15},
  number={7},
  pages={e2023MS003800},
  year={2023},
  publisher={Wiley Online Library}
}

@article{sachindra2018statistical,
  title={Statistical downscaling of precipitation using machine learning techniques},
  author={Sachindra, DA and Ahmed, Khandakar and Rashid, Md Mamunur and Shahid, S and Perera, BJC},
  journal={Atmospheric research},
  volume={212},
  pages={240--258},
  year={2018},
  publisher={Elsevier}
}

@article{li2020fourier,
  title={Fourier neural operator for parametric partial differential equations},
  author={Li, Zongyi and Kovachki, Nikola and Azizzadenesheli, Kamyar and Liu, Burigede and Bhattacharya, Kaushik and Stuart, Andrew and Anandkumar, Anima},
  journal={arXiv preprint arXiv:2010.08895},
  year={2020}
}

@misc{ncar_gdex_dataset_d633006,
  author = " {European Centre for Medium-Range Weather Forecasts}",
  title = {{ERA5 Reanalysis Model Level Data}},
  publisher = "NSF National Center for Atmospheric Research",
  address = "Boulder, CO",
  year = 2022,
  doi = "10.5065/XV5R-5344",
  url = "https://doi.org/10.5065/XV5R-5344"
}

@misc{guan_2026_18627189,
  author       = {Guan, Haiwen and
                  Darman, Moein and
                  Chakraborty, Dibyajyoti and
                  Arcomano, Troy and
                  Chattopadhyay, Ashesh and
                  Maulik, Romit},
  title        = {High-Resolution Climate Projections Using
                   Diffusion-Based Downscaling of a Lightweight
                   Climate Emulator
                  },
  month        = feb,
  year         = 2026,
  publisher    = {Zenodo},
  doi          = {10.5281/zenodo.18627189},
  url          = {https://doi.org/10.5281/zenodo.18627189},
}

@article{wan2023debias,
  title={Debias coarsely, sample conditionally: Statistical downscaling through optimal transport and probabilistic diffusion models},
  author={Wan, Zhong Yi and Baptista, Ricardo and Boral, Anudhyan and Chen, Yi-Fan and Anderson, John and Sha, Fei and Zepeda-N{\'u}{\~n}ez, Leonardo},
  journal={Advances in Neural Information Processing Systems},
  volume={36},
  pages={47749--47763},
  year={2023}
}

@article{wan2024regional,
  title={Regional climate risk assessment from climate models using probabilistic machine learning},
  author={Wan, Zhong Yi and Lopez-Gomez, Ignacio and Carver, Robert and Schneider, Tapio and Anderson, John and Sha, Fei and Zepeda-N{\'u}{\~n}ez, Leonardo},
  journal={arXiv preprint arXiv:2412.08079},
  year={2024}
}

@article{hess2025fast,
  title={Fast, scale-adaptive and uncertainty-aware downscaling of Earth system model fields with generative machine learning},
  author={Hess, Philipp and Aich, Michael and Pan, Baoxiang and Boers, Niklas},
  journal={Nature Machine Intelligence},
  volume={7},
  number={3},
  pages={363--373},
  year={2025},
  publisher={Nature Publishing Group UK London}
}

@article{aich2026conditional,
  title={Conditional diffusion models for downscaling and bias correction of Earth system model precipitation},
  author={Aich, Michael and Hess, Philipp and Pan, Baoxiang and Bathiany, Sebastian and Huang, Yu and Boers, Niklas},
  journal={Geoscientific Model Development},
  volume={19},
  number={4},
  pages={1791--1808},
  year={2026},
  publisher={Copernicus Publications G{\"o}ttingen, Germany}
}

@article{schmidt2025generative,
  title={A generative framework for probabilistic, spatiotemporally coherent downscaling of climate simulation},
  author={Schmidt, Jonathan and Schmidt, Luca and Strnad, Felix M and Ludwig, Nicole and Hennig, Philipp},
  journal={npj Climate and Atmospheric Science},
  volume={8},
  number={1},
  pages={270},
  year={2025},
  publisher={Nature Publishing Group UK London}
}

@article{ruhling2024probablistic,
  title={Probablistic Emulation of a Global Climate Model with Spherical DYffusion},
  author={R{\"u}hling Cachay, Salva and Henn, Brian and Watt-Meyer, Oliver and Bretherton, Christopher S and Yu, Rose},
  journal={Advances in Neural Information Processing Systems},
  volume={37},
  pages={127610--127644},
  year={2024}
}

@article{perkins2025hiro,
  title={HiRO-ACE: Fast and skillful AI emulation and downscaling trained on a 3 km global storm-resolving model},
  author={Perkins, W Andre and Kwa, Anna and McGibbon, Jeremy and Arcomano, Troy and Clark, Spencer K and Watt-Meyer, Oliver and Bretherton, Christopher S and Harris, Lucas M},
  journal={arXiv preprint arXiv:2512.18224},
  year={2025}
}

@article{watt2025ace2,
  title={ACE2: accurately learning subseasonal to decadal atmospheric variability and forced responses},
  author={Watt-Meyer, Oliver and Henn, Brian and McGibbon, Jeremy and Clark, Spencer K and Kwa, Anna and Perkins, W Andre and Wu, Elynn and Harris, Lucas and Bretherton, Christopher S},
  journal={npj Climate and Atmospheric Science},
  volume={8},
  number={1},
  pages={205},
  year={2025},
  publisher={Nature Publishing Group UK London}
}

@article{harris2022generative,
  title={A generative deep learning approach to stochastic downscaling of precipitation forecasts},
  author={Harris, Lucy and McRae, Andrew TT and Chantry, Matthew and Dueben, Peter D and Palmer, Tim N},
  journal={Journal of Advances in Modeling Earth Systems},
  volume={14},
  number={10},
  pages={e2022MS003120},
  year={2022},
  publisher={Wiley Online Library}
}

@inproceedings{price2022increasing,
  title={Increasing the accuracy and resolution of precipitation forecasts using deep generative models},
  author={Price, Ilan and Rasp, Stephan},
  booktitle={International conference on artificial intelligence and statistics},
  pages={10555--10571},
  year={2022},
  organization={PMLR}
}

@article{schillinger2025enscale,
  title={EnScale: Temporally-consistent multivariate generative downscaling via proper scoring rules},
  author={Schillinger, Maybritt and Samarin, Maxim and Shen, Xinwei and Knutti, Reto and Meinshausen, Nicolai},
  journal={arXiv preprint arXiv:2509.26258},
  year={2025}
}

@article{gelaro2017modern,
  title={The modern-era retrospective analysis for research and applications, version 2 (MERRA-2)},
  author={Gelaro, Ronald and McCarty, Will and Su{\'a}rez, Max J and Todling, Ricardo and Molod, Andrea and Takacs, Lawrence and Randles, Cynthia A and Darmenov, Anton and Bosilovich, Michael G and Reichle, Rolf and others},
  journal={Journal of climate},
  volume={30},
  number={14},
  pages={5419--5454},
  year={2017}
}

\appendix
\section{Hyperparameters}\label{hyperparams}
For EDM training, the noise level is sampled as $\log \sigma \sim \mathcal{N}\!\left(P_{\mathrm{mean}},\, P_{\mathrm{std}}^2\right)$. These hyperparameters were tuned empirically by testing on a smaller subset of the training data. Some of these are taken directly from \citet{karras2022elucidating} and \citet{chakraborty2026multimodal}. 

\begin{table}[!ht]
\centering
\small
\renewcommand{\arraystretch}{1.1}
\begin{tabularx}{\linewidth}{@{}l l X@{}}
\toprule
Symbol & Value & Description \\
\midrule
\multicolumn{3}{@{}l}{\textbf{Training}} \\
\midrule
$P_{\mathrm{mean}}$ & $0$ & Mean of the log-noise distribution used to sample training noise levels. \\
$P_{\mathrm{std}}$ & $2$ & Standard deviation of the log-noise distribution; larger values cover a wider range of noise scales. \\
$\sigma_{\mathrm{data}}$ & $1.2/1.0$ & 1.2 used for EDM loss weighting; 1.0 used inside the EDM preconditioner. \\
$\sigma_{\min}$ & $0$ & Lower bound of the noise level supported by the EDM preconditioner (a training-side preconditioner setting; the nonzero sampling endpoint $\sigma_{\min}=0.001$ in the Sampling block below is a distinct quantity, namely the final step of the sampling schedule). \\
$\sigma_{\max}$ & $80$ & Maximum noise level supported by the EDM preconditioner. \\
$\eta$ & $5\times10^{-5}$ & Learning rate. \\
 & $50000$ & Total training kilo-images (kimg)(refer to pairs of 6-hourly coarse and fine climate states) \\
 & $5000$ & Learning-rate warmup/ramp length in kimg. \\
 & $0.05$ & Minimum learning-rate factor at the end of decay. \\
\midrule
\multicolumn{3}{@{}l}{\textbf{Sampling}} \\
\midrule
$N$ & $35$ & Number of sampling steps. \\
$\sigma_{\min}$ & $0.001$ & Final noise level in the EDM sampling schedule. \\
$\sigma_{\max}$ & $80$ & Effective initial noise level used in sampling. The sampling routine passes a nominal request of $200$, but it is capped at the preconditioner's supported maximum of $80$; the effective value used throughout is therefore $80$, matching the training $\sigma_{\max}$, so this is a deliberate cap rather than a misconfiguration. \\
$\rho$ & $7$ & EDM schedule curvature controlling how steps are distributed across noise levels. \\
$S_{\mathrm{churn}}$ & $200$ & Strength of temporary stochastic noise injection during sampling. \\
$S_{\min}$ & $0.01$ & Lower bound of the noise range where churn is applied. \\
$S_{\max}$ & $20$ & Upper bound of the noise range where churn is applied. \\
$S_{\mathrm{noise}}$ & $1$ & Scale of the injected stochastic noise. \\
\bottomrule
\end{tabularx}
\vspace{0.1cm}
\caption{Common training and sampling hyperparameters for the diffusion-based super-resolution models.}
\label{tab:edm_hyperparameters}
\end{table}

\end{document}